\documentclass{article}

\usepackage{arxiv}

\usepackage{times}  
\usepackage{helvet}  
\usepackage{courier}  
\usepackage[hyphens]{url}  
\usepackage{graphicx} 
\usepackage{caption} 
\DeclareCaptionStyle{ruled}{labelfont=normalfont,labelsep=colon,strut=off} 
\usepackage{algorithm}
\usepackage{algorithmic}
\usepackage[utf8]{inputenc} 
\usepackage{url}            
\usepackage{amsmath}
\usepackage{booktabs}       
\usepackage{amsfonts}       
\usepackage{nicefrac}       
\usepackage{microtype}      
\usepackage{lipsum}		
\usepackage{graphicx}
\usepackage{doi}
\usepackage{amsthm}
\usepackage{subcaption}
\usepackage{pgfplots}
\usepackage{cancel}
\usepackage{tabularx}
\usepackage{multirow}
\usepackage{hhline}

\usepackage{newfloat}
\usepackage{listings}

\pgfplotsset{compat=1.18}

\theoremstyle{definition}
\newtheorem{definition}{Definition}

\theoremstyle{plain}

\newtheorem{theorem}{Theorem}
\newtheorem{lemma}{Lemma}

\DeclareMathOperator{\diag}{diag}
\DeclareMathOperator{\Var}{Var}

\DeclareMathOperator*{\argmin}{arg\,min}
\DeclareMathOperator{\Cov}{Cov}
\DeclareMathOperator*{\veco}{vec}

\usepackage{ifthen}
\usepackage{bigints}

\newcommand{\absof}[1]{\left|#1\right|}
\newcommand{\normof}[2]{\absof{\absof{#1}}_{#2}}
\newcommand{\squareof}[1]{\left(#1\right)^2}

\newcommand{\meanof}[1]{\mathbb{E}\left[{#1}\right]}
\newcommand{\meanofwrt}[2]{\mathbb{E}_{#2}\left[{#1}\right]}

\newcommand{\intreals}{\int_\mathbb{R}}

\newcommand{\biggintreals}{\bigintss_{\mathbb{R}}}

\newcommand{\expof}[1]{\exp\left({#1}\right)}
\newcommand{\gaussfrac}[1]{
	\ifthenelse{\equal{#1}{}}{\frac{1}{\sqrt{2\pi}}}{\frac{1}{\sqrt{2\pi{#1}}}}
}
\newcommand{\gaussexp}[2]{\expof{-\frac{1}{2}\frac{{#1}}{#2}}}

\newcommand{\expstdnorm}{\expof{-\frac{1}{2}x^2}}
\newcommand{\gaussdens}[2]{\gaussfrac{#2}\gaussexp{#1}{#2}}

\newcommand{\fatx}{\mathbf{x}}
\newcommand{\fatz}{\mathbf{z}}

\newcommand{\faty}{\mathbf{y}}
\newcommand{\fatX}{\mathbf{X}}

\newcommand{\fatm}{\mathbf{m}}
\newcommand{\fatf}{\mathbf{f}}
\newcommand{\fatF}{\mathbf{F}}
\newcommand{\fatK}{\mathbf{K}}
\newcommand{\fatk}{\mathbf{k}}
\newcommand{\fata}{\mathbf{a}}
\newcommand{\fatS}{\mathbf{S}}
\newcommand{\fatW}{\mathbf{W}}

\newcommand{\fatzero}{\mathbf{0}}
\newcommand{\fxb}{\fatx^B}
\newcommand{\Jk}{J_{k}}
\newcommand{\fatL}{\mathbf{L}}
\newcommand{\fatI}{\mathbf{I}}
\newcommand{\fatLambda}{\boldsymbol{\Lambda}}
\newcommand{\fatSigma}{\boldsymbol{\Sigma}}
\newcommand{\fatsigma}{\boldsymbol{\sigma}}
\newcommand{\fatmu}{\boldsymbol{\mu}}
\newcommand{\fatKNN}{\fatK_{NN}}

\newcommand{\mrfatfU}{m_{(r)|\fatf_U}}
\newcommand{\mrpkpfatfU}{m_{(r)|\fatf_U}^{\partial_k}}

\newcommand{\krfatfU}{k_{(r)|\fatf_U}}
\newcommand{\krpkpfatfU}{k_{(r)|\fatf_U}^{\partial_k}}
\newcommand{\krppkpfatfU}{k_{(r)|\fatf_U}^{\partial_k\partial_k'}}
\newcommand{\frackkppk}{\frac{\krpkpfatfU}{\krppkpfatfU}}

\newcommand{\fmufx}{\boldsymbol{\mu}_{\fatx}}

\newcommand{\fipk}[1]{f_{#1}^{\partial_k}}
\newcommand{\Si}[1]{S_{#1}}

\newcommand\Tstrut{\rule{0pt}{2.6ex}}       
\newcommand\Bstrut{\rule[-1.6ex]{0pt}{0pt}} 
\newcommand{\TBstrut}{\Tstrut\Bstrut} 

\usepackage[round]{natbib}

\usepackage[labelformat=simple, format=plain, width=0.95\textwidth]{caption}

\usepackage{float}

\title{Integrated Gradient attribution for Gaussian Processes with non-Gaussian likelihoods}

\author{ Sarem Seitz \\
  University of Bamberg\\
  \texttt{sarem.seitz@uni-bamberg.de}
  }

\begin{document}
\maketitle
\begin{abstract}
	Gaussian Process (GP) models are a powerful tool in probabilistic machine learning with a solid theoretical foundation. Thanks to current advances, modeling complex data with GPs is becoming increasingly feasible, which makes them an interesting alternative to deep learning and related approaches. As the latter are getting more and more influential on society, the need for making a model's decision making process transparent and explainable is now a major focus of research. A major direction in interpretable machine learning is the use of gradient-based approaches, such as Integrated Gradients, to quantify feature attribution, locally for a given datapoint of interest. Since GPs and the behavior of their partial derivatives are well studied and straightforward to derive, studying gradient-based explainability for GPs is a promising direction of research. Unfortunately, partial derivatives for GPs become less trivial to handle when dealing with non-Gaussian target data as in classification or more sophisticated regression problems. This paper therefore proposes an approach for applying Integrated Gradient-based explainability to non-Gaussian GP models, offering both analytical and approximate solutions. This extends gradient-based explainability to probabilistic models with complex likelihoods to extend their practical applicability.
\end{abstract}

\section{Introduction}
As machine learning is increasingly applied in critical domains such as healthcare and finance, the need for transparent models and algorithms has grown significantly. Although black-box models achieve high performance, their decision-making processes often remain opaque, raising concerns about consistency and trust. Issues related to fairness, representativeness, and safety necessitate improved model transparency, particularly in high-stakes applications. As a result, explainable machine learning is now an important and active domain of research.

An effective approach in this field is gradient-based explainability. These methods use partial derivatives of a model’s output with respect to its inputs to quantify feature relevance. Although this restricts the range of machine learning models that benefit from this approach to end-to-end differentiable ones, many of today's popular algorithms fulfill this requirement. 

Unlike typical black-box models, Gaussian Processes (GPs), inherently provide a certain degree of transparency, particularly through their kernel functions. Many popular kernel functions for GPs induce well-known functional properties for the resulting process. By selecting an appropriate covariance kernel, GPs can provide a level of transparency that is sometimes sufficient for interpretability. Consider here, for example, Additive Gaussian Processes as considered by \cite{duvenaud2011additive} and, recently, \cite{zhang2024gaussian}. 

\textbf{Problem.} For complex, highly non-linear problems with interactions between multiple input features, kernel-based transparency may still be insufficient, however. Based on a set of desirable properties for exlainability approaches, \cite{axiomaticintegratedgradients} propose Integrated Gradients to make generic, differentiable black-box models transparent and quantify the contribution of each input feature to a given model output. The authors show that Integrated Gradients fulfill all of the suggested requirements and are therefore particularly well aligned with what a human user would expect from explanations for black-box model predictions.

\textbf{Contribution.} Given the theoretical and practical amenities of attribution via Integrated Gradients, this paper extends their usage to GP models, namely those with non-Gaussian likelihood. We take advantage of the well-known property of GPs being closed under linear operations such as differentiation. In fact, \cite{butleriggp} derived the exact GP kernel of the theoretical Integrated Gradient formula for standard GP regression with Gaussian likelihood. This is remarkable insofar as the integral in Integrated Gradients usually has to be approximated and cannot be calculated in closed form. 
Here, we will also rely on an approximation of the integral but derive analytical formulas for non-Gaussian GP models using squared, probit and exponential inverse link-functions. For sigmoid and softmax inverse link-functions we derive approximate solutions via Taylor expansions. An additional by-product of the respective derivations is an analytical formula that allows to use quadrature methods to efficiently approximate the Integrated Gradient for arbitrary, single-output likelihoods. The proposed method is based on Sparse Variational GPs (SVGPs) but can be extended to any model that transforms a GP by some inverse link-function. 

\textbf{Related work.} In general, GPs offer intrinsic transparency through their kernel-based structure, making them distinct from many black-box models. This section reviews key methods for GP interpretability, including inducing points, Automatic Relevance Determination (ARD), and recent gradient-based approaches. The prototypical perspective is particularly relevant for variational GPs where so-called \emph{inducing points} - sometimes termed 'pseudo inputs' \citep{snelson2006sparse} - serve as compressed representations of the training data. After training the model, the learned inducing points often show prototypical patterns that are relevant for prediction. This is demonstrated in a particularly vivid manner in \cite{van2017convolutional} where a convolutional GP model learns the relevant patches of image classes as inducing features. Further research on variational and sparse variational GPs includes, amongst others, \cite{titsiassvgp, hensmanbig, hensman2015scalable}.

In contrast, the \emph{automatic relevance determination} (ARD) mechanism \citep[p. 106 ff.]{tipping2001sparse, williams2006gaussian} for kernel functions provides another way to achieve transparency in GP models. ARD kernels treat the detection of relevant input features as an optimization or bayesian inference problem where a weight vector determines the importance of each variable. The non-linearity of the kernel function then allows to model complex problems in that manner while, at the same time, being able to quantify the importance of a given input variable. In the past, the idea of ARD for feature importance has been relatively successful \citep{nipsfeaturecompetition}. However, the feature importances in ARD kernels are fixed, global values which can be too restrictive for complex data domains. This issue has more recently been addressed in \cite{de2024thin}, who propose an input dependent variant of the ARD kernel.

While ARD provides global feature relevance, kernel-based transparency offers a more flexible, functional approach. The choice of kernel function directly shapes the GP’s prior distribution over functions, influencing its posterior interpretability \citep{scholkopf2002learning, williams2006gaussian}. For example, a periodic kernel function implies a prior distribution over periodic functions and ultimately a posterior distribution over periodic functions. While this property can help to incorporate prior knowledge and infer general properties of GP posteriors, this inherent transparency does not help for our use-case. On another note, \cite{duvenaud2011additive} demonstrate how to use additive kernel functions in a GAM-like manner. In fact, their approach allows to infer the local influence of each input feature, which potentially allows to explain models with arbitrary input dimensionality. However, in order for this solution to yield interpretable results, we need to presume an additive structure among the transformed, individual input features. This is clearly a highly limiting assumption that might not yield satisfactory results for complex data either. 
Derivative centered inference in GPs itself is nothing new and has been advanced for example in \cite{solak2002derivative}, \cite{williams2006gaussian}, \cite{eriksson2018scaling} and \cite{de2021high}. Existing work that focuses primarily on derivative GPs in the context of gradient-based explanations, however, is not known to the author as per writing this paper. On the other hand, as is commonly known, derivatives and gradients can be used for local linear approximations of a target function. In the context of GPs, local approximations for GPs have been proposed by \cite{yoshikawa2021gaussian} via varying coefficient models.

Recently, Integrated Gradients in the context of GP regression with Gaussian likelihood has been extensively studied in \cite{butleriggp}, who derive the distributional parameters of the GP implied by applying Integrated Gradients to GP regression. \cite{zhang2025interpret} on the other hand propse a an approximation to Integrated Gradients by replacing the involved partial derivatives by infinitesimal differences. The results of \cite{chau2023explaining} show that non-gradient based explainability approaches like Shapley values also benefit from the properties of GPs, insofar as they allow to derive meaningful theoretical and practical results.

\textbf{Outline.} The remainder of this paper is structured as follows: We begin with a brief overview over the most important aspects of GP models, namely standard ones and the Sparse Variational GP (SVGP) approach. Each section also includes the corresponding partial derivative GPs for which analytical formulas exist for both cases. 
Thereafter, we discuss Integrated Gradients for explaining black-box models by creating feature attributions. Both concepts, SVGPs and Integrated gradients, are then merged to introduce Integrated Gradients for SVGPs, particularly for non-Gaussian likelihoods. The experimental part then provides evidence for the soundness and correctness of the proposed method. Finally, we look at potential limitations and possible future research. 

\textbf{Notation.} Throughout this paper, plain lower-case variables (e.g. $x$) denote scalar values, bold lower-case variables (e.g. $\fatx$) denote vectors, and bold upper-case variables (e.g. $\fatX$) denote matrices. The distinction whether the respective variables are deterministic or random will either be clear from context or explicity stated. Parentheses subscripts are used to denote specific elements of a variable, i.e. $\fatx_{(j)}$ is the $j$-th element of $\fatx$ and $\fatX_{(ij)}$ the element at the $i$-th row and $j$-th column. $\fatX_{(i\cdot)}$ denotes the $i$-th row of $\fatX$ and $\fatX_{(\cdot j)}$ the $j$-th column. A full list of additionally introduced, relevant symbols can be found in Appendix A.

\section{Gaussian Processes and sparse variational approximations}

In this section, we briefly review the main types of GP models that are relevant to us. These methods are standard \emph{Gaussian Process} regression models and \emph{Sparse Variational Gaussian Processes} (SVGPs).

\subsection{Gaussian Process Regression}
The building blocks of GPs, see \citet{rasmussen2003gaussian}, are a prior distribution over functions, $p(f)$, and a likelihood $p(y|f)$. Using Bayes' law, we are interested in a posterior distribution $p(f|y)$ obtained as 
\begin{equation}\label{bayeslaw}
    p(f|y)=\frac{p(y|f)p(f)}{p(y)}.
\end{equation}
The prior distribution is a Gaussian Process, fully specified by a mean function $m(\cdot):\mathcal{X}\mapsto\mathbb{R}$, typically $m(x)=0$,  and covariance kernel function $k(\cdot,\cdot):\mathcal{X}\times\mathcal{X}\mapsto \mathbb{R}_0^+$:
\begin{equation}\label{gaussproc}
    f\sim\mathcal{GP}(m(\cdot),k(\cdot,\cdot))
\end{equation}
A common choice for $k(\cdot,\cdot)$ is the Squared exponential (SE) kernel
\begin{equation}\label{ardkern}
    k_{SE}(\fatx,\fatx')=\theta\cdot exp\left(-0.5 \frac{||\fatx-\fatx'||^2}{l^2}\right)
\end{equation}
where $l\in\mathbb{R}$ and $\theta>0$. Evaluation of $f$ at a finite set of inputs, here an $N\times M$ row-matrix $\fatX_N$, yields a multivariate Gaussian distribution over function outputs:
\begin{equation}
    f(\fatX_N)=:\fatf_N\sim \mathcal{N}(\fatm_N,\fatKNN)
\end{equation}
where $\fatm_N$ denotes the evaluation of $m(\cdot)$ at each row of $\fatX_N$ and $\fatK_{NN}$ denotes the positive semi-definite \textit{Gram-Matrix}, obtained as $\fatK_{NN(ij)}=k(\fatx_i,\fatx_j)$, $\fatx_i$ the $i$-th row of  $\fatX_N$. From now on, we will match the finite dimensional evaluations of GPs, mean functions and kernel functions to their respective inputs via matching subscript indices - i.e. a GP $f$ evaluated at input matrix $\fatX_A$ will be denoted as $\fatf_A$. Kernel gram-matrices between two differing input sets, say $\fatX_A$ and $\fatX_B$, will be denoted as $\fatK_{AB}$.

We now presume that $m(\fatx)=0$ for the remainder of this paper, without losing generality as all results are straightforward to extend to arbitrary, differentiable mean functions. Provided that $p(\faty_N|\fatf_N)=\prod_{i=1}^N \mathcal{N}(\faty_{(i)}|\fatf_{(i)},\sigma^2)$, i.e. observations are i.i.d. univariate Gaussian conditional on $\fatf_N$, it is possible to directly calculate a corresponding posterior distribution for a matrix of unseen inputs, $\fatX_*$, as
\begin{equation}\label{postpred}
	p(\fatf_*|\faty_N)=\mathcal{N}(\fatf_*|\tilde{\fatLambda}_{*N} \faty_N,\fatK_{**}-\tilde{\fatLambda}_{*N} (\fatK_{NN}+I\sigma^2) \tilde{\fatLambda}_{*N}^T)
\end{equation}
where $\tilde{\fatLambda}_{*N}=\fatK_{*N}(\fatK_{NN}+\fatI_N\sigma^2)^{-1}$. $\fatI_N$ is the $N\times N$ identity matrix. Throughout the rest of this paper, we will set $m(x)=0$. 

To optimize model hyperparameters, either maximum likelihood or Markov-Chain Monte-Carlo optimization are the most popular techniques. Both approaches, however, don't scale well for larger datasets and a 'naive' implementation of GP regression will quickly become computationally infeasbile. As a result, a broad range of scalable GP methods has been developed over the years.

Similarly to \eqref{postpred}, for GPs with differentiable mean-function and twice-differentiable kernel-function \citep{adler2010geometry}, one can calculate a posterior distribution for derivative observations $\fatf^{\partial_k}_*=\frac{\partial}{\partial\fatx_k}f(\fatX_*)$:
\begin{equation}\label{postpredderiv}
	p(\fatf^{\partial_k}_*|\faty_N)=\mathcal{N}(\fatf^{\partial_k}_*|\tilde{\fatLambda}^{\partial_k}_{*N} \faty,\fatK^{\partial_k\partial_k'}_{**}-\tilde{\fatLambda}^{\partial_k}_{*N} (\fatK_{NN}+I\sigma^2) \tilde{\fatLambda}_{N*}^{\partial_k'})
\end{equation}
Here, $\tilde{\fatLambda}^{\partial_k}_{*N}=\fatK^{\partial_k}_{*N}(\fatK_{NN}+\fatI_N\sigma^2)^{-1}$, $\tilde{\fatLambda}^{\partial_k'}_{N*}=(\fatK_{NN}+\fatI_N\sigma^2)^{-1}\fatK^{\partial_k'}_{N*}$ and $\fatK^{\partial_k}_{*N}$, $\fatK^{\partial_k'}_{N*}$ denote the evaluation at $\fatX_*,\fatX_N$ of the partial derivative kernel functions $k^{\partial_k}(\fatx,\fatx')=\frac{\partial}{\partial\fatx_k}k(\fatx,\fatx')$, $k^{\partial_k'}(\fatx,\fatx')=\frac{\partial}{\partial\fatx_k'}k(\fatx,\fatx')$. Also, $\fatK^{\partial_k\partial_k'}_{**}$ denotes the evaluation at $\fatX_*$ of the partial derivative kernel in both arguments, $k^{\partial_k\partial_k'}(\fatx,\fatx')=\frac{\partial^2}{\partial\fatx_k\partial\fatx_k'}k(\fatx,\fatx')$. 

While GP models with Gaussian likelihood yields the above, closed-form posterior model, introducing another likelihood to account for other distributions of observed data usually results in an intractable posterior. Both, the scalability issue and the problem of intractable posteriors are often being accounted for by Sparse Variational Gaussian Proccesses (\emph{SVGP}s).

\subsection{Sparse Variational Gaussian Processes and non-Gaussian likelihoods}
Sparse variational inference for GPs, both for Gaussian and non-Gaussian likelihoods, was formally developed in \cite{hensmanbig, hensmansvgpc}, but also builds upon the seminal work of \cite{titsiassvgp}. SVGPs introduce a set of $U$ so called inducing locations $\fatX_U\subset \mathcal{X}$ and corresponding inducing variables $\fatf_U$, where $U<<N$. A respective posterior distribution, $p(f,\fatf_U|\faty_N)$, is then approximated through a variational distribution $q(f,\fatf_U)=p(f|\fatf_U)q(\fatf_U)$ - usually $q(\fatf_U)=\mathcal{N}(\fatf_U|\fata,\fatS),\fatS=\fatL \fatL^T$ - by maximizing the \emph{evidence lower bound} (ELBO) with respect to model parameters, inducing locations and inducing variable:
\begin{equation}\label{standardelbo}
	ELBO = \sum_{i=1}^N\mathbb{E}_{p(f|\fatf_U)q(\fatf_U)}\left[\log p(\faty_{N(i)}|g(\fatf_{N(i)}))\right]-KL(q(\fatf_U)||p(\fatf_U))
\end{equation}
where we obtain $p(\fatf_U)$ by evaluating the GP prior distribution at inducing locations $\fatX_U$. Here, $g(\cdot)$ denotes an \emph{inverse link-function} to map the GP to a domain suitable for the target likelihood-parameter. Common choices for $g(\cdot)$ are, for example, the probit function or the sigmoid function for binary classification problems and the exponential function for likelhoods involving a parameter with positive range. Notice that we omit potential additional likelihood-parameters in our notation for simplicity.

Following standard results for Gaussian random variables, it can also be shown that, for marginal $\fatf_*$, evaluated at an arbitrary input matrix $\fatX_*$ and with $\fatLambda_{*U}=\fatK_{*U}(\fatK_{UU}-\fatS)^{-1}$, we have
\begin{equation}\label{marginalvariationalfunction}
	q(\fatf_*)=\int p(\fatf_*|\fatf_U)q(\fatf_U)d\fatf_U=\mathcal{N}(\fatf_*|\fatK_{*U}\fatK_{UU}^{-1} \fata,\fatK_{**}-\fatLambda_{*U} (\fatK_{UU}-\fatS)\fatLambda_{*U}^T)
\end{equation}
Notice that \eqref{marginalvariationalfunction} implies the existence of a corresponding variational posterior GP with mean and kernel functions,
\begin{equation}\label{varpostgpparams}
	m_{|\fatf_U}(\fatx)=k(\fatx,\fatX_U)\fatK_{UU}^{-1}\fata \quad\text{and}\quad		k_{|\fatf_U}(\fatx,\fatx')=k(\fatx,\fatx')-k(\fatx,\fatX_U)\fatK_{UU}^{-1}(\fatK_{UU}-\fatS)\fatK_{UU}^{-1}k(\fatX_U,\fatx'),
\end{equation}
where $k(\fatx,\fatX_U)$ and $k(\fatX_U,\fatx)$ denote the kernel function evaluation for $\fatx$ against all elements of $\fatX_U$. Evaluation of \eqref{varpostgpparams} at $\fatx_*$ is consequently denoted as, $m_{*|\fatf_U}$ and $k_{*|\fatf_U}$ evaluation at $\fatX_*$ is denoted as $\fatm_{*|\fatf_U}$ and $\fatK_{**|\fatf_U}$.

To handle problems with $C$ output dimensions, we introduce $C$ statistically independent GPs to model the variational posterior and extend \eqref{standardelbo} as follows:
\begin{equation}\label{elbomultioutput}
	ELBO = \sum_{i=1}^N\mathbb{E}_{p(\fatF|\fatF_U)q(\fatF_U)}\left[\log p(\faty_{N(i)}|G\left(\fatF_{N(i\cdot)}\right)\right]-\sum_{j=1}^C KL(q(\fatf^{(j)}_U)||p(\fatf^{(j)}_U))
\end{equation}
Here, $F$ denotes stacking of the individual GPs, $f^{(1)},...,f^{(C)}$; $\fatF_{N(i\cdot)}$ denotes stacking the $i$-th elements of each individual GP, evaluated at $\fatX_N$. We can think of $\fatF_{N}$ a column-matrix whose $j$-th column denotes the evaluation $f^{(j)}(\fatX_N)$, which makes $\fatF_{N(i\cdot)}$ the $i$-th row of $\fatF_N$. Notice that all elements in $\fatF_N$ follow a matrix-variate Gaussian distribution. Given independence of the individual GPs, the columns of $\fatF_N$ must be independent, too, for arbitrary $\fatX_N$.

$G(\cdot)$ is now a multivariate inverse link-function that preserves dimensionality. Our interest here is primarily the softmax inverse link-function but other alternatives are possible, depending on the problem at hand.

The challenge in any non-Gaussian SVGP ELBO optimization is typically intractability of the expected log-likelihood term. While the an analytical solution is sometimes possible, one often resorts to Gauss-Hermite quadrature to approximate the solution for single-GP models and Monte-Carlo approximation for multi-GP models.  

Before continuing, we introduce the following Lemma that provides an analogous result for GP derivative observations in the SVGP setup:

\begin{lemma}\label{svgpderivobs}

	Let $k(\cdot,\cdot)$ be a twice-differentiable kernel function, $f\sim\mathcal{GP}(0,k(\cdot,\cdot))$ denote a GP and $f^{\partial_k}\sim\mathcal{GP}(0,k^{\partial_k\partial_k'}(\cdot,\cdot))$ its partial derivative process with respect to the $k$-th input. Also, let $q(\fatf_U)=\mathcal{N}(\fatf_U|\fata,\fatS)$ as above. For $q(\fatf_*, \fatf^{\partial_k}_*)=\int p(\fatf_*,\fatf^{\partial_k}_*|\fatf_U)q(\fatf_U)d\fatf_U$ it then follows that
\begin{equation}\label{marginalvariationalderivfunction}
	q(\fatf_*,\fatf^{\partial_k}_*)=\mathcal{N}\left(\begin{bmatrix}\fatf_* \\ \fatf^{\partial_k}_*\end{bmatrix}\bigg|\begin{bmatrix}\fatK_{*U}\fatK_{UU}^{-1}\fata \\ \fatK^{\partial_k}_{*U}\fatK_{UU}^{-1}\fata\end{bmatrix} ,
\begin{bmatrix}\fatK_{**}-\fatLambda_{*U} (\fatK_{UU}-\fatS)\fatLambda_{*U}^T & \fatK^{\partial_k'}_{**}-\fatLambda_{*U} (\fatK_{UU}-\fatS)\fatLambda_{U*}^{\partial_k'}\\ 
\fatK^{\partial_k}_{**}-\fatLambda_{*U}^{\partial_k} (\fatK_{UU}-\fatS)\fatLambda_{U*} & \fatK^{\partial_k\partial_k'}_{**}-\fatLambda^{\partial_k}_{*U} (\fatK_{UU}-\fatS)\fatLambda_{U*}^{\partial_k'} \end{bmatrix}\right)
\end{equation}
where $\fatLambda_{*U}^{\partial_k}=\fatK^{\partial_k}_{*U}(\fatK_{UU}-\fatS)^{-1}$ and $\fatLambda_{U*}^{\partial_k'}=(\fatK_{UU}-\fatS)^{-1}\fatK^{\partial_k'}_{U*}$.
\end{lemma}

A proof can be found in Appendix A. According to \eqref{marginalvariationalderivfunction}, we can easily obtain derivative function evaluations for the variational posterior by plugging in the corresponding kernel evaluations into \eqref{marginalvariationalfunction}. This will turn out to be essential when calculating Integrated Gradients for SVGPs later on.

Similarly to \eqref{varpostgpparams}, the implicit mean and kernel functions of the variational posterior derivative GP process in \eqref{marginalvariationalderivfunction} are
\begin{equation}\label{varpostderivgpparams}
	m^{\partial_k}_{|\fatf_U}(\fatx)=k^{\partial_k}(\fatx,\fatX_U)\fatK_{UU}^{-1}\fata \quad\text{and}\quad		k^{\partial_k\partial_k'}_{|\fatf_U}(\fatx,\fatx')=k^{\partial_k\partial_k'}(\fatx,\fatx')-k^{\partial_k}(\fatx,\fatX_U)\fatK_{UU}^{-1}(\fatK_{UU}-\fatS)\fatK_{UU}^{-1}k^{\partial_k'}(\fatX_U,\fatx).
\end{equation}
In addition, \eqref{marginalvariationalderivfunction} also implies a cross-covariance kernel to model the covariance between $f(\fatx)$ and $f^{\partial_k}(\fatx')$ under the variational posterior $q(f(\fatx),f^{\partial_k}(\fatx'))$:
\begin{equation}\label{varpostderivgpcrosscov}
	\begin{split}
		k^{\partial_k'}_{|\fatf_U}(\fatx,\fatx')&=k^{\partial_k'}(\fatx,\fatx')-k(\fatx,\fatX_U)\fatK_{UU}^{-1}(\fatK_{UU}-\fatS)\fatK_{UU}^{-1}k^{\partial_k'}(\fatX_U,\fatx')\\
		&=:\meanofwrt{\Cov(f(\fatx),f^{\partial_k}(\fatx')|\fatf_U)}{q(\fatf_U)}
	\end{split}
\end{equation}
By symmetry of the covariance operator, the equivalence
\begin{equation}\label{varpostderivgpcrosscov2}
	k^{\partial_k'}_{|\fatf_U}(\fatx,\fatx')=k^{\partial_k}_{|\fatf_U}(\fatx',\fatx)=k^{\partial_k}(\fatx',\fatx)-k^{\partial_k}(\fatx',\fatX_U)\fatK_{UU}^{-1}(\fatK_{UU}-\fatS)\fatK_{UU}^{-1}k(\fatX_U,\fatx)\\
\end{equation}
follows.

\section{Post-hoc, gradient-based attribution}
A popular approach to explain the predictions of black-box models are attributions for which \cite{axiomaticintegratedgradients} provides a formal definition:

\begin{definition}\label{attrdef}
An attribution of the prediction of a function $f(\cdot)$ at input $\fatx\in\mathbb{R}^M$ relative to a baseline input $\fatx^B\in\mathbb{R}^N$ is a vector $A_f(\fatx,\fatx^B)=(a_1,...,a_M)\in\mathbb{R}^N$ where $a_j$ is the contribution of $x_j$ to the prediction $f(\fatx)$.
\end{definition}

Presume an almost everywhere differentiable model function, $f$. A popular way to generate an attribution vector to locally, at given input $\fatx$ explain the model's decision, is the element-wise product of the input vector with the model function's gradient with respect to the input:
\begin{equation}\label{naiveattr}
A_f(\fatx,\fatx^B) = \fatx\odot\nabla f(\fatx)
\end{equation}
This approach is based on the idea of a locally linear approximation of the target function around $\fatx$. Besides not having a clear way to incorporate the baseline $\fatx^B$, this approach can yield attributions that go against human intution. \cite{layerwiserelevance, axiomaticintegratedgradients} identify several properties that are required for an attribution method to be well aligned with human intuition and show that attribution via Integrated Gradients and path-based attribution methods in general satisfy all of them. These properties are \emph{linearity}, \emph{sensitivity}, \emph{implementation invariance} and \emph{completeness}, where the latter implies that, for a given input, the sum of all attributions is equal to the difference between the model output given that input and given the baseline:
\begin{equation}\label{igdiffproperty}
\sum_{j=1}^M a_j=f(\fatx)-f(\fatx^B)
\end{equation}
The definition of an Integrated Gradient for the $i$-th feature of a given input vector $\fatx$ for model $f$ is as follows:
\begin{equation}\label{igdefinition}
IG(\fatx,\fatx^B)^f_k=(\fatx_k-\fatx^B_k)\times \int_{0}^1 \frac{\partial f(\fatx^B+\alpha\times(\fatx-\fatx^B))}{\partial\fatx_k} d\alpha
\end{equation}
A closed form solution for the line integral in \eqref{igdefinition} is often impossible for more complex models. The alternative is to approximate the integral by a Riemann sum instead: 
\begin{equation}\label{igapprox}
\hat{IG}(\fatx,\fatx^B)^f_k=\frac{\fatx_k-\fatx^B_k}{R}\sum_{r=1}^R \frac{\partial f(\fatx^B+\frac{r}{R}\times(\fatx-\fatx^B))}{\partial\fatx_k}
\end{equation}
For large $R$, we therefore have $\hat{IG}(\fatx,\fatx^B)_k\approx IG(\fatx,\fatx^B)_k$

The baseline $\fatx^B$ can be chosen based on the given problem domain for which model explanations should be generated. Quite often, the choice is $\fatx^B=\boldsymbol{0}$ which further simplifies the above calculation. From now, we will focus on the approximate variant \eqref{igapprox} - a discussion of this and similar decisions will be done at the end.

\section{Gradient-based explanations for Gaussian Processes with non-Gaussian likelihood}
We now want to apply Integrated Gradient attribution to the case where $f$ is a single- or multi-output GP or, particularly for the non-Gaussian case, its element-wise transformation by an inverse-link function. Here, we are particulary interested in the variational posterior GP, after fitting its parameters to the data. Our targets for generating attributions is therefore 
\begin{equation}\label{multioutput}
	\begin{gathered}
	G_c(F(\cdot)):=G(f^{(1)}(\cdot),...,f^{(C)}(\cdot))_{(c)}, \\
		f^{(j)}(\cdot)\sim\mathcal{GP}(m_{|\fatf^{(j)}_U}^{(j)}(\cdot), k_{|\fatf^{(j)}_U}^{(j)}(\cdot,\cdot))
	\end{gathered}
\end{equation}
to explain the c-th prediction of a multi-output, non-Gaussian SVGP. Notice that \eqref{multioutput} encapsulates single-output models by replacing $G_c$ by $g$ and considering a single GP $f$ with corresponding partial derivative processes. From now, we will only consider the single-output case explicitly when necessary.
In the case of classfication, where $g$ could be the probit or sigmoid and $G$ the softmax inverse link-function, the attributions from \eqref{multioutput} would then quantify the contribution of each feature towards the predicted class-probabilities. 
In the general case, analogously to Generalized Linear Models, $g(f(\cdot))$ can be used to model an arbitrary, meaningful summary statistic of the target likelihood, conditioned on $\fatx$ -  typically the conditional mean or the variance. Corresponding attributions then explain the contribution of each feature towards the model predicting the respective summary statistic.
Applying the Integrated Gradient approximation in \eqref{igapprox} to \eqref{multioutput} directly yields
\begin{equation}\label{iggpmultiv}
\hat{IG}(\fatx,\fxb)_k^{G_c\circ F}=\frac{\fatx_k-\fxb_k}{R}\sum_{r=1}^R \nabla G_c(F(\fatx_{(r)}))^T\cdot F^{\Jk}(\fatx_{(r)})
\end{equation}
where $\fatx_{(r)}:=\fxb+\frac{r}{R}\times(\fatx-\fxb)$ and the elements of the multi-output GP $F^{\Jk}$ are the partial derivative processes (with respect to $\fatx_k$) corresponding to the elements of $F$\footnote{Let $JF$ denote the Jacobian of $F$ and we are interested in its $k$-th column, which we denote as $J_k F$. Denoting the vector of corresponding partial derivative GPs as $F^{\Jk}$ is then consistent with the previous notation for partial derivative GPs in the single-output case}.    
When $G_c(\cdot)$ is the identity functions as in standard SVGP regression with (multivariate) Gaussian likelihood, \eqref{iggpmultiv} can easily be shown to yield another GP with closed-form mean and kernel functions, due to linearity of the expectation operation. The case for general $G_c(\cdot)$ and different likelihoods is, obviously, far less convenient.
Since the (approximated) posterior mean is the typical point predictor in SVGP models, we consequently put our focus on generating attributions for 
\begin{equation}\label{meanssvgp}
	\meanofwrt{G_c(F(\fatx))}{q(F(\fatx))}.
\end{equation}
By pulling the respective partial derivative operators into the mean, it is trivial to show that the mean of the Integrated Gradient processes and the Integrated Gradients for \eqref{meanssvgp} coincide. For the multi-output case, we proceed similarly and explicitly decompose the gradient dot-product into a sum of partial derivatives:
\begin{equation}\label{meanofigmulti}
	\begin{gathered}
		\meanofwrt{\hat{IG}(\fatx,\fxb)_k^{G_c\circ F}}{q(\fatF(\fatx_{(1...R)}),\fatF^{\Jk}(\fatx_{(1...R)}))} = \frac{\fatx_k-\fxb_k}{R}\sum_{r=1}^R \meanofwrt{\nabla G_c(F(\fatx_{(r)}))^T\cdot F^{\Jk}(\fatx_{(r)})}{q(\fatF(\fatx_{(r)}),\fatF^{\Jk}(\fatx_{(r)}))}\\
		=\frac{\fatx_k-\fxb_k}{R}\sum_{r=1}^R \sum_{j=1}^C\meanofwrt{\frac{\partial}{\partial f^{(j)}}G_c(f^{(1)}(\fatx_{(r)}),...,f^{(j)}(\fatx_{(r)})) f^{(r)\partial_k}(\fatx_{(r)})}{q(\fatF(\fatx_{(r)}),\fatF^{\Jk}(\fatx_{(r)}))}
	\end{gathered}
\end{equation}
By linearity of the expectation operation, this reduces the problem of finding the mean Integrated Gradient to finding analytical or approximate solutions for
\begin{equation}\label{meanofigsummulti}
	\meanofwrt{\frac{\partial}{\partial f^{(j)}}G_c(f^{(1)}(\fatx_{(r)}),...,f^{(j)}(\fatx_{(r)})) f^{(r)\partial_k}(\fatx_{(r)})}{q(\fatF(\fatx_{(r)}),\fatF^{\Jk}(\fatx_{(r)}))}.
\end{equation}
Plugging a result for \eqref{meanofigsummulti} into \eqref{meanofigmulti} then yields a solution.

The following Lemma is key to the remaining results and can trivially be proven via the law of total expectation\footnote{Write $\meanof{h(\fatx)y}=\meanof{h(\fatx)\meanof{y|\fatx}}$ and put in the formula for the conditional Gaussian mean.}:

\begin{lemma}\label{proplawtotalexp}
	Let $\bar{\fatx}$ denote a $C+1$-dimensional Gaussian random vector. Fix $\fatx$ and $y$, where $y$ is an arbitrary element of $\bar{\fatx}$ and $\fatx$ denotes the remaining elements of $\bar{\fatx}$ without $y$. Denote by $\mu_{\fatx},\mu_y$ the mean vector and mean scalar of $\fatx$ and $y$, by $\Sigma_{\fatx}$ the covariance matrix of $\fatx$ and by $\Sigma_{y\fatx}$ the transposed vector of covariances between $y$ and each element of $\fatx$.

Then, for a function $h:\mathbb{R}^C\mapsto A\subseteq\mathbb{R}$, with $h$ continuous, it follows that
$$\mathbb{E}\left[h(\fatx)\cdot y\right]=\mathbb{E}\left[h(\fatx)\cdot\left(\mu_y+
\Sigma_{y\fatx}\Sigma_{\fatx}^{-1}(\fatx-\mu_\fatx)\right)\right].$$
\end{lemma}

Besides being key to the main result of this paper, Lemma \ref{proplawtotalexp} can directly be used to efficiently approximate the Integrated Gradient for single-output models with arbitrary inverse link-function via Gauss-Hermite quadrature. For specific single-output inverse link-functions and the softmax, multi-output function, we derive the following main Theorem:

\begin{theorem}
	Let $q(f)$ or $q(f^{(1)},...,f^{(R)})$ be SVGP variational posterior distributions of a single or multi-output and $g(\cdot)$ or $G(\cdot)$ non-linear inverse link-functions whose first derivative is well-defined almost everywhere; in particular, let $g(\cdot)$ be either the square function, the natural exponential function, the Probit function $\Phi(\cdot)$ or the sigmoid function $s(cdot)$ and let $G(\cdot)$ be the softmax function. The corresponding mean Integrated Gradient attributions for a corresponding SVGP model as described above can then be calculated as follows: 

    \begin{tabularx}{\textwidth}{l|X}
        \hline
		\small\textbf{Predictive mean} & \small\textbf{Integrated Gradient mean} $\meanof{\hat{IG}(\fatx,\fxb)_k^{g\circ f}}$ \\
        \hline\hline
		\small $\meanofwrt{f(\fatx)^2}{q(f(\fatx))}$ & \small $=\frac{\fatx_k-\fxb_k}{R}\sum_{r=1}^R \left(2\mrfatfU\mrpkpfatfU+2\krpkpfatfU\right)$ \TBstrut \\
		\small $\meanofwrt{\expof{f(\fatx)}}{q(f(\fatx))}$ &  \small $=\frac{\fatx_k-\fxb_k}{R}\sum_{r=1}^R\expof{\mrfatfU+0.5\krfatfU}\cdot\left(\mrpkpfatfU+\frackkppk\krfatfU\right)$  \TBstrut \\
		\small $\meanofwrt{\Phi(f(\fatx))}{q(f(\fatx))} $ & \small $=\frac{\fatx_k-\fxb_k}{R}\sum_{r=1}^R\frac{1}{\sqrt{2\pi(1+\krfatfU)}}\gaussexp{\mrfatfU^2}{1+\krfatfU}\cdot\left(\mrpkpfatfU-\frackkppk\frac{\mrfatfU\krfatfU}{1+\krfatfU}\right)$ \TBstrut \\
		\small $\meanofwrt{s(f(\fatx))}{q(f(\fatx)}$ & \small $\approx \frac{\fatx_k-\fxb_k}{R}\sum_{r=1}^R\left(s'(\mrfatfU)\mrpkpfatfU + \frac{1}{2}s'''(\mrfatfU)\mrpkpfatfU + \frackkppk s''(\mrfatfU)\right)$
 \TBstrut \\
		\small $\meanofwrt{S_c(F(\fatx))}{q(F(\fatx))}$ & \small$\approx\begin{gathered} \frac{(\fatx_k-\fxb_k)}{R}\sum_{r=1}^R \left(\vphantom{\sum_{j\in\{1,...,C\}\setminus c}}H^k_{c(r)}(\fatm_{(r)|\fatf_U})+0.5\diag(\fatK_{(r)|\fatf_U})^T\diag(\nabla^2 H^k_{c(r)}(\fatm_{(r)|\fatf_U}))\right. \\\quad\quad\left.- \sum_{j\in\{1,...,C\}\setminus c} H_{cj(r)}^k(\fatm_{(r)|\fatf_U})+0.5\diag(\fatK_{(r)|\fatf_U})^T\diag(\nabla^2 H_{cj(r)}^k(\fatm_{(r)|\fatf_U}))\right)\end{gathered}$ \TBstrut \\
        \hline
    \end{tabularx}

where the exact definitions of $H^k_{c(r)}(\fatz)$ and $H^k_{cj(r)}(\fatz)$ for the softmax model can be found in Appendix C.5.
\end{theorem}

The approximate solutions for the sigmoid and softmax cases are derived via second-order Taylor expansion around the means of $f(\fatx)$ and $f^{(j)}(\fatx)$. Unfortunately, the complexity of the result makes it challenging to derive a reasonably tight error bound and is therefore left for future research.  

Applying the fundamental theorem for line integrals to the above, similarly to the deterministic case, we get the next Theorem that is applicable to a sufficiently wide range of inverse link-functions, including the ones discussed in this paper. A proof can be found in Appendix D.

\begin{theorem}\label{convergencetheorem}
	Let $F:\mathbb{R}^M\mapsto\mathbb{R}^W$ denote a vector-valued GP and $G_c(\cdot)$ the $c$-th output of a corresponding multivariate inverse-link function $G:\mathbb{R}^W\mapsto\mathbb{R}^C$, such that $G_c$ has continuous partial derivatives almost everyhwere. Also, denote by $F^{\Jk}(\cdot)$ the $k$-th column of the Jacobian $JF(\cdot)$. 
If either
	\begin{itemize}
		\item $\sup_{j\in 1,...,C}\left(|\nabla G_c(\faty)|\right)_{(j)}\leq D\left(1+\left(\sum_{j=1}^W|\faty_{(j)}|\right)^r\right),\quad D>0,r\geq 0$ or
		\item $\sup_{j\in 1,...,C}\left(|\nabla G_c(\faty)|\right)_{(j)}\leq \kappa\expof{\sum_{(j)=1}^W\gamma_j \fatx_{(j)}},\quad \kappa>0, \gamma_j\geq0$
	\end{itemize}
	then

	\begin{equation}
		\sum_{k=1}^{M}\meanofwrt{\hat{IG}(\fatx,\fxb)_k^{G_c\circ F}}{q(F(\fatx_{(1...R)},F^{\Jk}(\fatx_{(1...R)})))} \rightarrow \meanofwrt{G_c(F(\fatx))}{q(F(\fatx))}-\meanofwrt{G_c(F(\fxb))}{q(F(\fxb))}
	\end{equation}
	as $R\rightarrow \infty$. 
\end{theorem}
Notice that Theorem 2 also covers the case of single-output GPs and corresponding inverse link-functions. For sufficiently large $R$, we then use
\begin{equation}\label{igdiffs}
	\left|\sum_{k=1}^{M}\meanofwrt{\hat{IG}(\fatx,\fxb)_k^{g\circ f}}{q(f(\fatx_{(1...R)},f^{\partial_k}(\fatx_{(1...R)})))} - \left(\meanofwrt{g(f(\fatx))}{q(f(\fatx))}-\meanofwrt{g(f(\fxb))}{q(f(\fxb))}\right)\right|
\end{equation}
as a proxy metric to evaluate the exactness of our approximations, both with respect to the empirical Integrated Gradient and with respect to the approximate solutions for sigmoid and softmax inverse links. A reasonably 
exact approximation should have small error and thus fulfill the completeness condition of Integrated Gradients rather well.

\section{Experiments}
In this section, we quantitatively evaluate the proposed method. We begin by validating Theorem 2 for both the exact and approximate formulas on synthetic datasets. Furthermore, the proposed method is evaluated on real-world datasets. For all experiments, the expected model output is either calculated analytically or approximated via Gauss-Hermite quadrature (sigmoid inverse link-function) or Monte-Carlo integration (softmax inverse link-function). Further experiment-details can be found in Appendix E and the documentation of the GPyTorch Python library \citep{gardner2018gpytorch}, which has been used primarily for this section.
\subsection{Convergence of Empirical Integrated Gradients on synthetic data}
In conjunction with Theorem \ref{convergencetheorem}, we test if \eqref{igdiffs} does indeed reduce when we increase the number of points in the Riemann sum-approximation for the corresponding Integrated Gradients. For the exact details on the data generation and the evaluation itself, the reader is referred to Appendix E.1.
\begin{figure}[htbp!]
\begin{center}
\begin{tabularx}{0.675\textwidth}{cl|c|c|c|c} 
    \hline
    && \multicolumn{4}{c}{\textbf{\# Riemann Points}} \\ 
	&Inverse link-function& 50 & 500 & 1000 & 5000 \\
   \hhline{==|=|=|=|=}
	\parbox[t]{2mm}{\multirow{3}{*}{\rotatebox[origin=c]{90}{Exact}}} & Square & 0.0014 & 0.0001 & 0.0001 & 0.0000 \\
    & Exponential &  0.0078 & 0.0009 & 0.0004 & 0.0001 \\ 
    & Probit &  0.0008 & 0.0001 & 0.0001 & 0.0000 \\ 
   \hhline{==|=|=|=|=}
	\parbox[t]{2mm}{\multirow{3}{*}{\rotatebox[origin=c]{90}{Appr.}}} & Sigmoid (Gauss-Hermite appr.) & 0.0006 & 0.0001 & 0.0000 & 0.0000 \\
    & Sigmoid (Taylor appr.) & 0.0007 & 0.0004 & 0.0004 & 0.0004 \\ 
    & Softmax (Taylor appr.) &  0.0013 & 0.0003 & 0.0002 & 0.0002 \\ 
    \hline
\end{tabularx}
	\caption{\textit{Absolute difference between sum of (approximated) Integrated Gradients and the difference between expected model output for target and baseline input.}}
	\label{riemannsumtable}
\end{center}
\end{figure}
Figure \ref{riemannsumtable} provides evidence that, for large number of approximation points, the \emph{empirical} Integrated Gradient attributions indeed converges to the \emph{theoretical} Integrated Gradients. For the exact variants in the top half of the evaluation table, this merely indicates implementation correctness as any other behavior would directly contradict Theorem \ref{convergencetheorem}. For the approximate fomulas, i.e. for Taylor approximation with sigmoid and softmax inverse link-function, this also gives some evidence that the additional approximation error from the Taylor-series expansion is not too large to prohibit practical applicability.
\subsection{Application on real-world datasets}
Here, we look at exemplary Integrated Gradient attributions two real-world datasets:

\textbf{Bike-sharing demand} \citep{bikeshare} - Poisson likelihood, exponential inverse link-function. 

\begin{figure}[htbp!]
    \centering
    \begin{minipage}{0.6\linewidth}
        \centering
        \medskip
        \begin{subfigure}[t]{.9\linewidth}
            \centering
            \includegraphics[width=.99\linewidth]{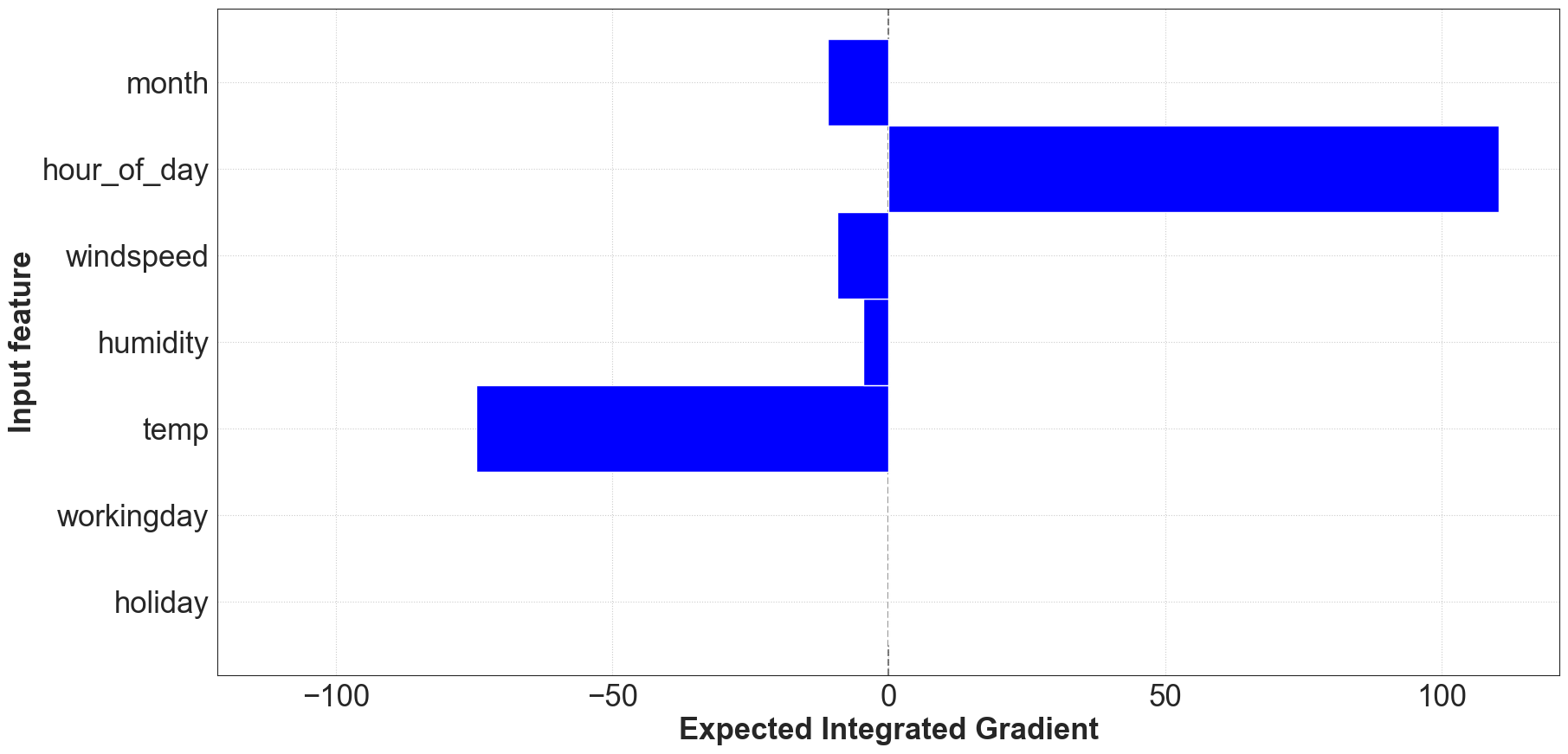}
        \end{subfigure}
        \label{fig:bikeshareig}
    \end{minipage}
    \begin{minipage}{0.35\linewidth}
        \centering
        \begin{tabularx}{\linewidth}{l|X|X}
            \hline
             & \textbf{Target} & \textbf{Baseline}\\
            \hline\hline
            month & 1 & 6 \\
            hour\_of\_day & 9 & 12 \\
            windspeed & 20.00 & 12.80 \\
			humidity & 37.00 & 61.89 \\
			temp & 9.02 & 20.23 \\
			workingday & 1 & 1 \\
			holiday & 0 & 0 \\
            \hline
        \end{tabularx}
    \end{minipage}
	\caption{\textit{Bike sharing demand - example Integrated Gradients (left) with target and baseline features (right)}}
    \label{fig:bikeresult}
\end{figure}

This dataset consists of several weather and datetime features over time to predict the count of bikes rented at a given point in time. Here, the existing features \emph{holiday}, \emph{workingday} (both 0-1 encoded), \emph{temp}, \emph{humidity},and \emph{windspeed} (numeric) are used. Two additional features are created from the \emph{datetime} variable: \emph{hour\_of\_day}, the given hour of the record (0-23), and \emph{month}, the given month of the record (1-12). Both features are additionally transformed into their harmonic representation,
\begin{equation}
	f_F(x)=\begin{pmatrix}\sin\left(\frac{2\pi}{F}x\right) & \cos\left(\frac{2\pi}{F}x\right)\end{pmatrix},
\end{equation}
where either $F=24$ or $F=12$. To derive the respective attribution for the original feature, we sum the attributions of both harmonic features. Finally, all features are divided by their maximum-value to ensure a common scale among them.
As a baseline for calculating the Integrated Gradient, we choose the training-sample average of all numeric features, \emph{temp}, \emph{humidity},and \emph{windspeed} and set $\text{holiday}=0,\,\text{workingday}=1,\,\text{hour\_of\_day}=12,\,\text{month}=6$. Qualitatively, our attribution therefore explains the output deviation for an observation with average weather conditions, on a workingday in June at noon.
Figure \ref{fig:bikeresult} shows the attribution for an example target under the above conditions.Notably, the attribution of the \emph{holiday} and \emph{workingday} features is zero, which is expected due to their equivalence to these features in the baseline example.

\textbf{MNIST and FashionMNIST} \citep{deng2012mnist, xiao2017fashion} - Multinomial likelihood, softmax inverse link-function; ablation experiments. 
\begin{figure}[htbp!]
  \centering
  \begin{subfigure}[t]{.45\linewidth}
    \centering
    \begin{minipage}{.5\linewidth}
		\raisebox{0.45cm}{
		  \includegraphics[width=\linewidth]{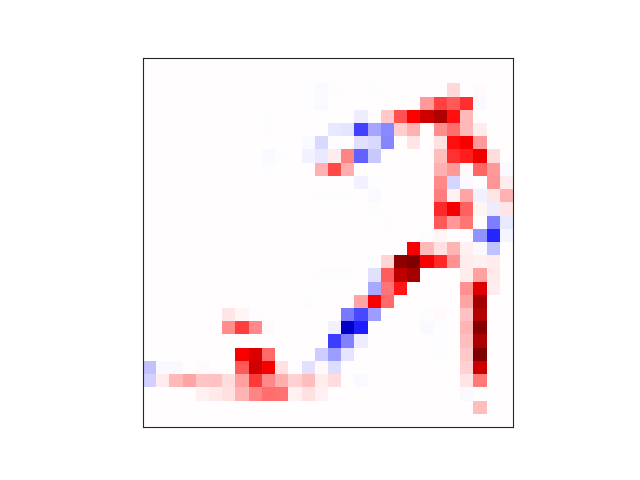}
		}
    \end{minipage}%
    \begin{minipage}{.5\linewidth}
		\includegraphics[width=\linewidth]{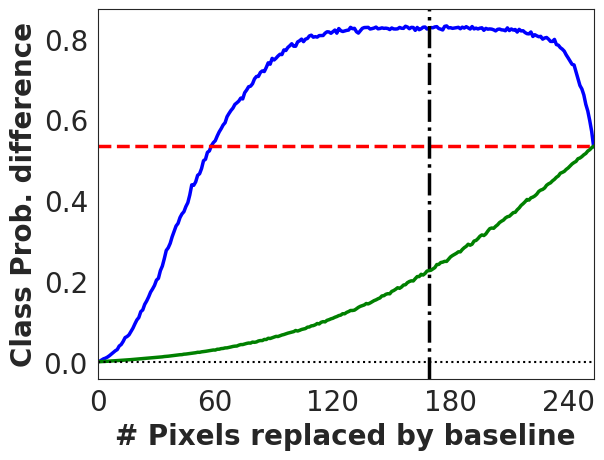}
    \end{minipage}
  \end{subfigure}
  \begin{subfigure}[t]{.45\linewidth}
    \centering
    \begin{minipage}{.5\linewidth}
		\raisebox{0.45cm}{
		  \includegraphics[width=\linewidth]{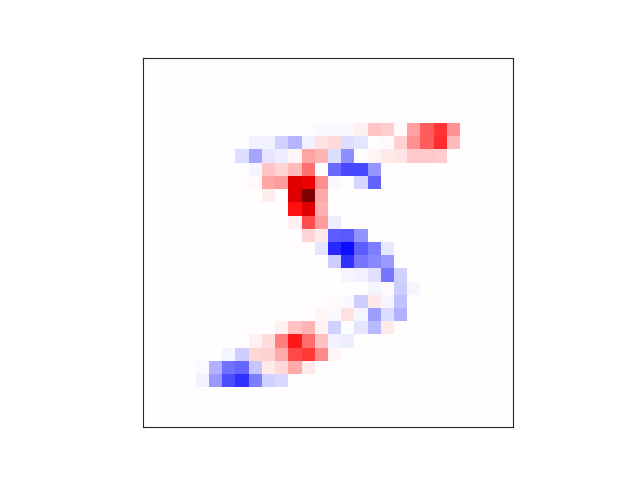}
		}
    \end{minipage}%
    \begin{minipage}{.5\linewidth}
      \includegraphics[width=\linewidth]{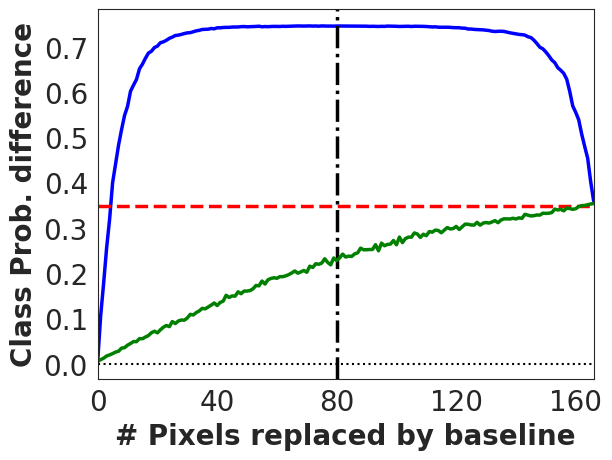}
    \end{minipage}
  \end{subfigure}

	\caption{\textit{MNIST/FashionMNIST ablation examples. The left image of each subfigure depicts Integrated Gradient attribution (red=positive attribution; blue=negative attribution). Right image of each subfigure shows how class probability changes when pixels are replaced by their baseline counterpart, either in order of their Integrated Gradient attribution (\textbf{blue line}) or at random (\textbf{green line}, averaged over 1000 trials).}}
  \label{fig:mnist_plot_small}
\end{figure}

To verify feasibility of the proposed method for small-sized image data, we apply it on the MNIST and FasionMNIST datasets. As pointed out in \cite{axiomaticintegratedgradients}, the theoretical properties of Integrated Gradients already guarantee sensible attributions. However, since our calculations of Integrated Gradients for softmax inverse link-function relies on Taylor-approximation, the coherence of the resulting attributions could suffer from approximation error. To find out if this is the case, an ablation experiment similar to \cite{samek2016evaluating} is performed - a detailed rationale for the graphs can be found in Appendix E.2.

In short, we compare the model's predicted class probability for the correct target class, given input $\fatx$, against the predicted class probability of a perturbed version of the image, $\tilde{\fatx}$, as follows:
\begin{equation}\label{outputprobdiff}
	\meanof{S_c(F(\fatx))}-\meanof{S_c(F(\tilde{\fatx}))}
\end{equation}
The perturbed images are built iteratively, where, at each iteration, one more pixel is replaced by its baseline counterpart, $\fatx^B$, here the all-black image where all pixels are encoded as zeros. In Figure \ref{fig:mnist_plot_small}, the \emph{blue} line represents the difference \eqref{outputprobdiff} when performing this replacement in order of the Integrated Gradient attributions, from most positive to most negative attribution value. The \emph{green} line shows, at each iteration, this difference when dropping the same number of pixels at random, averaged over 1000 draws. 

Since, for sufficiently large $R$, Integrated Gradient attributions explain the difference in output between target input and baseline input (Theorem 2),
\begin{equation}\label{baselineprobdiff}
	\sum_{k=1}^{M}\meanofwrt{\hat{IG}(\fatx,\fxb)_k^{S_c\circ F}}{q(F(\fatx_{(1...R)},F^{\Jk}(\fatx_{(1...R)})))} \approx \meanofwrt{S_c(F(\fatx))}{q(F(\fatx))}-\meanofwrt{S_c(F(\fxb))}{q(F(\fxb))},
\end{equation}
bringing $\tilde{\fatx}$ closer to $\fxb$ systematically based on attribution order, should result in steeper adjustments than doing so randomly. Indeed, the blue lines all show fairly steep increase at the beginning, when we replace features with large, positive Integrated Gradient attribution. The \emph{black}, vertical line depicts the feature, at which the ordered Integrated Gradients switch signs from positive to negative. According to our outlined rationale, we should see the predicted differences start decreasing slowly at first (features with negative attributions closer to zero are replaced first). At the end of the replacement process, we see a steep drop of the blue line as their corresponding attributions become increasingly negative. 

Finally, as $\tilde{\fatx}\equiv\fxb$, the output-difference roughly converges to the sum of integrated gradients (\emph{red} line) as predicted in \eqref{baselineprobdiff}. Notice that the convergence is not exact here, due to the approximation error caused by the Taylor approximation. These empirical observations show the coherence of the generated attributions and cannot be explained by pure coincidence, given that random feature replacement doesn't show such systematic behavior. Further examples of these kinds of ablation experiments can be found in Appendix E.3.

\section{Discussion}
This work has demonstrated that Integrated Gradients can be effectively extended to Gaussian Process models, particularly in handling non-Gaussian likelihoods, where we derived analytical and approximate solutions for different inverse link-functions. For inverse link-functions such as squared, probit, and exponential, we have derived analytical solutions for empirical Integrated Gradients. Sigmoid and softmax inverse link-functions, while lacking a closed-form solution, can still be handled approximately - either via Gauss-Hermite quadrature or via second-order Taylor approximation.

\textbf{Limitations.} Regarding limiations and room for improvement, we can identify three broad categories: \textit{Scalability -} While we were able to train the model on a high-dimensional and reasonably big dataset in regards to the general capabilities of GP models, real-world datasets are often even larger both in dimensionality and number of training examples. In particular high-dimensional image data as typical for real-world image problems can be difficult to handle, depending on practical latency requirements. As each new dimension requires $R$ additional calculations, this can become a hinderance. \textit{Approximation accuracy and guarantees -} For softmax inverse link-functions, the approximation remains an open challenge due to the nonlinearity of the transformation. Our empirical evaluations suggest that Taylor approximations introduce systematic bias in attributions. Future work should explore higher-order approximations or adaptive quadrature methods to mitigate this error. 
\textit{Attribution uncertainty -} Another notable drawback is that it focuses solely on the mean of the Integrated Gradient distribution, disregarding posterior variance. This means we cannot directly quantify the uncertainty of attributions, which could be important for model debugging and fairness considerations. One possible extension is to establish results for posterior variance or posterior uncertainy for feature attributions in general, which is left for future research. 

\textbf{Other directions.} Besides improving on the above weaknesses and limitations, research on GP explainability could shed further light on explainability for Deep Learning, too. Besides classical connections between GPs and Neural Networks as initially introduced in the seminal work of \cite{neal2012bayesian}, recent work has added the \textit{Neural Tangent Kernel} \citep{jacot2018neural} or efficient Laplace approximations for posterior distributions in Bayesian Deep Learning \citep{daxberger2021laplace}. These connections could establish a stronger theoretical foundation for respective interpretability approaches and improve their performance further.

\bibliographystyle{plainnat}  
\bibliography{references} 

\appendix
\newpage
\section{A. List of symbols and notation}
\begin{table}[htbp!]
    \centering
    \begin{tabularx}{\textwidth}{lX}
        \hline
        \textbf{Symbol} & \textbf{Explanation} \\
        \hline
        $\fatf_*$ & GP evaluated at matrix of input row-vectors $\fatX_*$ \\
	 	$\fatf_*^{\partial_k}$ & Partial derivative (w.r.t. $\fatx_k$) GP evaluated at matrix of input row-vectors $\fatX_*$ \\	
		$\fatf_U$ & GP evaluated at matrix of inducing row-vectors $\fatX_U$ \\
		$k^{\partial_k}(\cdot,\cdot)$ & Partial derivative (w.r.t $k$-th input feature) kernel (w.r.t. \textbf{first} kernel argument) \\
		$k^{\partial_k'}(\cdot,\cdot)$ & Partial derivative (w.r.t $k$-th input feature) kernel (w.r.t. \textbf{second} kernel argument) \\
		$k^{\partial_k\partial_k'}(\cdot,\cdot)$ & Partial derivative (w.r.t $k$-th input feature) kernel (w.r.t. \textbf{both} kernel arguments) \\
		$\fatK_{AB}$ & Kernel Gram-matrix, corresponding to kernel $k(\cdot,\cdot)$, evaluated at $\fatX_A$ as first argument \newline and $\fatX_B$ as second argument \\
		$\fatK_{AB}^{\partial_k}$ & Kernel Gram-matrix, corresponding to kernel $k^{\partial_k}(\cdot,\cdot)$, evaluated at $\fatX_A$ as first argument \newline and $\fatX_B$ as second argument \\
		$\fatK_{AB}^{\partial_k'}$ & Kernel Gram-matrix, corresponding to kernel $k^{\partial_k'}(\cdot,\cdot)$, evaluated at $\fatX_A$ as first argument \newline and $\fatX_B$ as second argument \\
		$\fatK_{AB}^{\partial_k\partial_k'}$ & Kernel Gram-matrix, corresponding to kernel $k^{\partial_k\partial_k'}(\cdot,\cdot)$, evaluated at $\fatX_A$ as first argument \newline and $\fatX_B$ as second argument \\
		$\fatLambda_{*U}$ & $:=\fatK_{*U}\fatK_{UU}^{-1}$ \\
		$\fatLambda_{*U}^{\partial_k}$ & $:=\fatK^{\partial_k}_{*U}\fatK_{UU}^{-1}$ \\
		$\fatLambda_{U*}^{\partial_k'}$ & $:=\fatK_{UU}^{-1}\fatK^{\partial_k'}_{U*}$ \\
		$\fatx_{(r)}$ & $:=\fxb+\frac{r}{R}\times(\fatx-\fxb)$ \\
		$k_{**}$ & Evaluation of kernel function $k(\cdot,\cdot)$ at a single input vector $\fatx_*$ \\
		$k^{\partial_k\partial_k'}_{**}$ & Evaluation of kernel function $k{\partial_k\partial_k'}(\cdot,\cdot)$ at a single input vector $\fatx_*$ \\
		$\fatk_{*U}$ & Evaluation row-vector corresponding to kernel function $k(\cdot,\cdot)$ at a single input vector $\fatx_*$ as first and input matrix $\fatX_U$ as second argument \\
		$\fatk^{\partial_k}_{*U}$ & Evaluation row-vector corresponding to kernel function $k^{\partial_k}(\cdot,\cdot)$ at a single input vector $\fatx_*$ as first and input matrix $\fatX_U$ as second argument \\
		$\fatk^{\partial_k'}_{U*}$ & Evaluation row-vector corresponding to kernel function $k^{\partial_k}(\cdot,\cdot)$ at an input matrix $\fatX_U$ as first and input vector $\fatx_*$ as second argument \\

		$m_{*|\fatf_U}$ & $:=\fatk_{*U}\fatK^{-1}_{UU}\fata$ \\
		$k_{*|\fatf_U}$ & $:=k_{**}-\fatk_{*U}\fatK^{-1}_{UU}(\fatK_{UU}-\fatS)\fatK^{-1}_{UU}\fatk_{U*}$\\
		$\fatm_{*|\fatf_U}$ & $:=\fatK_{*U}\fatK^{-1}_{UU}\fata$ \\
		$\fatK_{**|\fatf_U}$ & $:=\fatK_{**}-\fatK_{*U}\fatK^{-1}_{UU}(\fatK_{UU}-\fatS)\fatK^{-1}_{UU}\fatK_{U*}$\\
		$m^{\partial_k}_{*|\fatf_U}$ & $:=\fatk^{\partial_k}_{*U}\fatK^{-1}_{UU}\fata$ \\
		$k^{\partial_k\partial_k'}_{*|\fatf_U}$ & $:=k^{\partial_k\partial_k'}_{**}-\fatk^{\partial_k}_{*U}\fatK^{-1}_{UU}(\fatK_{UU}-\fatS)\fatK^{-1}_{UU}\fatk^{\partial_k'}_{U*}$\\
		$k^{\partial_k}_{*|\fatf_U}$ & $:=k^{\partial_k}_{**}-\fatk^{\partial_k}_{*U}\fatK^{-1}_{UU}(\fatK_{UU}-\fatS)\fatK^{-1}_{UU}\fatk_{U*}$\\
        \hline
    \end{tabularx}
    \caption{List of symbols used throughout the paper}
    \label{tab:symbols}
\end{table}

\newpage

\section{B. Proof of Lemma 1}
\begin{proof}[\unskip\nopunct]
We have for the joint distribution of $\fatf_*$, $\fatf_*^{\partial_k}$ and $\fatf_U$:
\begin{align*}
    p(\fatf_*,\fatf_*^{\partial_k},\fatf_U)=
    \mathcal{N}\left(\begin{bmatrix}
    \fatf_*\\
    \fatf_*^{\partial_k}\\
    \fatf_U
    \end{bmatrix}\Bigg\vert\begin{bmatrix}
	\mathbf{0}\\
	\mathbf{0}\\
    \fatf_U
    \end{bmatrix},\begin{bmatrix}
    \fatK_{**} & \fatK_{**}^{\partial_k'} & \fatK_{*U}\\
    \fatK_{**}^{\partial_k} & \fatK_{**}^{\partial_k\partial_k'} & \fatK_{*U}^{\partial_k} \\
    \fatK_{U*} & \fatK_{U*}^{\partial_k'} & \fatK_{UU}
    \end{bmatrix}\right)
\end{align*}
\noindent (notice that $m(\fatx)=0$ for all $x$ per our initial choice and therefore $m^{\partial_k}(\fatx)=0$ for all $x$ as well)\\
By symmetry of the covariance matrix we therefore have $\left(\fatK_{*U}^{\partial_k}\right)^T=\fatK_{U*}^{\partial_k'}$. Placing the variational distribution over $\fatf_U$, we get
\begin{align*}
q(\fatf_*, \fatf_*^{\partial_k})&=\int p(\fatf_*,\fatf_*^{\partial_k}|\fatf_U)q(\fatf_U)\,d\fatf_U\\
&=\int \mathcal{N}\left(\begin{bmatrix}\fatf_* \\ \fatf^{\partial_k}_*\end{bmatrix}\bigg|\begin{bmatrix}\fatK_{*U}\fatK_{UU}^{-1}\fatf_U \\ \fatK^{\partial_k}_{*U}\fatK_{UU}^{-1}\fatf_U\end{bmatrix} ,
	\begin{bmatrix}\fatK_{**}-\fatK_{*U} \fatK_{UU}^{-1}\fatK_{U*} & \fatK^{\partial_k'}_{**}-\fatK_{*U} \fatK_{UU}^{-1}\fatK_{U*}^{\partial_k'}\\ 
\fatK^{\partial_k}_{**}-\fatK_{*U}^{\partial_k} \fatK_{UU}^{-1}\fatK_{U*} & \fatK^{\partial_k\partial_k'}_{**}-\fatK^{\partial_k}_{*U} \fatK_{UU}^{-1}\fatK_{U*}^{\partial_k'} \end{bmatrix}\right)
 q(\fatf_U)\,d\fatf_U \\
	&=\mathcal{N}\left(\begin{bmatrix}\fatf_* \\ \fatf^{\partial_k}_*\end{bmatrix}\bigg|\begin{bmatrix}\fatK_{*U}\fatK_{UU}^{-1}\fata \\ \fatK^{\partial_k}_{*U}\fatK_{UU}^{-1}\fata\end{bmatrix} \vphantom{\begin{bmatrix}\left(\fatLambda_{*U}^{\partial_k}\right)^T \\ \left(\fatLambda_{*U}^{\partial_k}\right)^T \end{bmatrix}},\right. \\
	&\quad\quad\quad\quad\left.\begin{bmatrix}\fatK_{**}-\fatK_{*U} \fatK_{UU}^{-1}\fatK_{U*}+\fatLambda_{*U} \fatS\fatLambda_{*U}^T & \fatK^{\partial_k'}_{**}-\fatK_{*U} \fatK_{UU}^{-1}\fatK_{U*}^{\partial_k'} +\fatLambda_{*U} \fatS\left(\fatLambda_{*U}^{\partial_k}\right)^T\\ 
	\fatK^{\partial_k}_{**}-\fatK_{*U}^{\partial_k} \fatK_{UU}^{-1}\fatK_{U*}+\fatLambda_{*U}^{\partial_k} \fatS\fatLambda_{*U}^T & \fatK^{\partial_k\partial_k'}_{**}-\fatK^{\partial_k}_{*U} \fatK_{UU}^{-1}\fatK_{U*}^{\partial_k'}+\fatLambda^{\partial_k}_{*U} \fatS\left(\fatLambda_{*U}^{\partial_k}\right)^T \end{bmatrix}\right)\\
&=\mathcal{N}\left(\begin{bmatrix}\fatf_* \\ \fatf^{\partial_k}_*\end{bmatrix}\bigg|\begin{bmatrix}\fatK_{*U}\fatK_{UU}^{-1}\fata \\ \fatK^{\partial_k}_{*U}\fatK_{UU}^{-1}\fata\end{bmatrix} ,
\begin{bmatrix}\fatK_{**}-\fatLambda_{*U} (\fatK_{UU}-\fatS)\fatLambda_{*U}^T & \fatK^{\partial_k'}_{**}-\fatLambda_{*U} (\fatK_{UU}-\fatS)\fatLambda_{U*}^{\partial_k'}\\ 
\fatK^{\partial_k}_{**}-\fatLambda_{*U}^{\partial_k} (\fatK_{UU}-\fatS)\fatLambda_{U*} & \fatK^{\partial_k\partial_k'}_{**}-\fatLambda^{\partial_k}_{*U} (\fatK_{UU}-\fatS)\fatLambda_{U*}^{\partial_k'} \end{bmatrix}\right)
\end{align*}

\end{proof}

\section{C. Proof of Theorem 1}
\begin{proof}[\unskip\nopunct]
To prove this result, we recall \eqref{meanofigmulti} and then apply Lemma 2 to calculate or approximate, via second-order Taylor approximation, each mean summand in the corresponding Riemann sums individually. 

In the single output case,

\begin{align}\label{univiggptarget}
	\meanofwrt{\hat{IG}(\fatx,\fatx')_k^{g\circ f}}{q(f(\fatx_{(1...R)}),f^{\partial_k}(\fatx_{(1...R)}))} = (\fatx_k-\fatx_k') \times \frac{1}{R}\sum_{r=1}^R \meanofwrt{g'(f(\fatx_{(r)}))\cdot f^{\partial_k}(\fatx_{(r)})}{q(f(\fatx_{(r)}),f^{\partial_k}(\fatx_{(r)}))},
\end{align}

this is $\meanofwrt{g'(f(\fatx_{(r)}))\cdot f^{\partial_k}(\fatx_{(r)})}{q(f(\fatx_{(r)}))}$.

The multi-output case,

\begin{gather}\label{multiggptarget}
	\meanofwrt{\hat{IG}(\fatx,\fatx')_k^{G_c\circ F}}{q(\fatF(\fatx_{(1...R)}),\fatF^{\Jk}(\fatx_{(1...R)}))} \\
	  \frac{\fatx_k-\fatx_k'}{R}\sum_{r=1}^R \sum_{j=1}^C\meanofwrt{\frac{\partial}{\partial f_{(j)}}G_c(f^{(1)}(\fatx_{(r)}),...,f_{(j)}(\fatx_{(r)})) \cdot f^{(j)\partial_k}(\fatx_{(r)})}{q(\fatF(\fatx_{(r)}),\fatF^{\Jk}(\fatx_{(r)}))},
\end{gather}

is more involved as we have to handle the double sum, but still follows the same principle.

In the single-output case we are dealing with means of Gaussian products, $\meanof{h(\fatx)y}$, as required in Lemma 2. Here, $f(\fatx_{(r)})$ and $f^{\partial_k}(\fatx_{(r)})$ take the role of $\fatx$ and $y$; the corresponding bi-variate Gaussian is given by
\begin{equation}\label{univdists}
	q\left(\begin{bmatrix}f(\fatx_{(r)}) \\ f^{\partial_k}(\fatx_{(r)})\end{bmatrix}\right) = \mathcal{N}\left(\begin{bmatrix}f(\fatx_{(r)}) \\ f^{\partial_k}(\fatx_{(r)})\end{bmatrix}\bigg|
\begin{bmatrix}m_{(r)|\fatf_U} \\ m^{\partial_k}_{(r)|\fatf_U}\end{bmatrix}, \begin{bmatrix}
	k_{(r)|\fatf_U} & k^{\partial_k}_{(r)|\fatf_U} \\
		k^{\partial_k}_{(r)|\fatf_U} & k^{\partial_k\partial_k'}_{(r)|\fatf_U}\end{bmatrix}\right).
\end{equation}

To make the subsequent derivations more convenient to read, we will substitute the dense notation above by the bi-variate Gaussian with marginals $x,y$, derive our result and then retransform into the original notation to state the final result. I.e., we have

\begin{equation}\label{univdistsxy}
	q\left(\begin{bmatrix}f(\fatx_{(r)}) \\ f^{\partial_k}(\fatx_{(r)})\end{bmatrix}\right) \Rightarrow q\left(\begin{bmatrix}x \\ y\end{bmatrix}\right) = \mathcal{N}\left(\begin{bmatrix}x \\ y\end{bmatrix}\bigg|
\begin{bmatrix}\mu_x \\ \mu_y \end{bmatrix}, \begin{bmatrix}
	\sigma_x^2 & \sigma_{xy} \\
\sigma_{xy} & \sigma_y^2\end{bmatrix}\right).
\end{equation}

The different, univariate inverse link-functions then represent $h(\cdot)$. By exploiting the interaction of these functions with the Gaussian mean, we are then able to derive analytical solutions for the most popular inverse links. The rest can efficiently be approximated with Gaussian quadrature or Taylor approximation.

For the softmax case the approach is essentially the same but with the additional complexity of handling a multi-input function and multiple GPs at once. Hence, we will postpone further investigation of this case to the last subsection of this proof. 

The remainder of this proof is divided into five subsections, each deriving the respective results per inverse link-function. 

\subsection{C.1 Derivation for the square inverse link-function}
For squared inverse link-function, 
\begin{equation}
g(f(\fatx_{(r)}))=f(\fatx_{(r)})^2,
\end{equation}
the partial derivative for given $\fatx_{(r)}$ is
\begin{equation}
\frac{\partial}{\partial\fatx_k}g(f(\fatx_{(r)}))=2\cdot(f(\fatx_{(r)}))\cdot f^{\partial_k}(\fatx_{(r)}).
\end{equation}
For generic Gaussian distribution, we have
\begin{equation}
\begin{split}
\meanof{g'(x)y}&=2\meanof{xy} \\
&=2\meanof{x\left(\mu_y+\frac{\sigma_{xy}}{\sigma^2_x}\left(x-\mu_x\right)\right)} \\
&=2\left(\mu_y-\frac{\sigma_{xy}}{\sigma^2_x}\mu_x\right)\meanof{x}+2\frac{\sigma_{xy}}{\sigma^2_x}\meanof{x^2} \\
&=2\left(\mu_y-\frac{\sigma_{xy}}{\sigma^2_x}\mu_x\right)\mu_x+2\frac{\sigma_{xy}}{\sigma^2_x}\left(\mu_x^2+\sigma_x^2\right) \\
&=2\mu_x\mu_y+2\sigma_{xy}.
\end{split}
\end{equation}
Plugging back in the distributional parameters from \eqref{univdists}, we get
\begin{equation}\label{squaresolution}
\meanofwrt{g'(f(\fatx_{(r)}))\cdot f^{\partial_k}(\fatx_{(r)})}{q(f\fatx_{(r)},f^{\partial_k}(\fatx_{(r)}))}=2\mrfatfU\mrpkpfatfU+2\krpkpfatfU
\end{equation}

By plugging \eqref{squaresolution} into \eqref{meanofigmulti}, we get the proclaimed result:
\begin{equation}
\begin{split}
	\meanofwrt{\hat{IG}(\fatx,\fxb)_k^{g\circ f}}{q(f(\fatx_{(1...R)}),f^{\partial_k}(\fatx_{(1...R)}))} &= \frac{\fatx_k-\fxb_k}{R}\sum_{r=1}^R \meanofwrt{g'(f(\fatx_{(r)}))\cdot f^{\partial_k}(\fatx_{(r)})}{q(f(\fatx_{(r)}),f^{\partial_k}(\fatx_{(r)}))} \\
&=\frac{\fatx_k-\fxb_k}{R}\sum_{r=1}^R \left(2\mrfatfU\mrpkpfatfU+2\krpkpfatfU\right)
\end{split}
\end{equation}

\subsection{C.2 Derivation for the Probit inverse link-function}
For probit inverse link-function, 
\begin{equation}
g(f(\fatx_{(r)}))=\Phi\left(f(\fatx_{(r)})\right),
\end{equation}
the partial derivative for given $\fatx_{(r)}$ is
\begin{equation}
\frac{\partial}{\partial\fatx_k}g(f(\fatx_{(r)}))=\phi(f(\fatx_{(r)}))\cdot f^{\partial_k}(\fatx_{(r)}).
\end{equation}
where $\phi(\cdot)$ denotes the standard Gaussian density function.
For generic Gaussian distribution, we have
\begin{equation}
	\begin{split}
\meanof{g'(x)y}&=\meanof{\phi(x)y} \\
&=\meanof{\phi(x)\left(\mu_y+\frac{\sigma_{xy}}{\sigma^2_y}\left(x-\mu_x\right)\right)} \\
&=\left(\mu_y-\frac{\sigma_{xy}}{\sigma^2_y}\mu_x\right)\meanof{\phi(x)}+\frac{\sigma_{xy}}{\sigma^2_y}\meanof{\phi(x)x}
\end{split}
\end{equation}
\begin{equation}
	\begin{split}
\meanof{\phi(x)}&=\intreals \gaussfrac{} \expstdnorm \gaussdens{(x-\mu_x)^2}{\sigma_x^2} dx \\
&=\frac{1}{2\pi\sigma_x}\intreals \gaussexp{x^2-2x\mu_x+\mu_x^2+\sigma_x^2x^2}{\sigma_x^2} dx \\
&=\frac{1}{2\pi\sigma_x}\intreals \gaussexp{(\sigma_x^2+1)x^2-2x\mu_x+\mu_x^2}{\sigma_x^2} dx \\
&=\frac{1}{2\pi\sigma_x}\cdot\\
&\cdot\biggintreals\expof{-\frac{(1+\sigma_x^2)x^2-(1+\sigma_x^2)2x\frac{\mu}{1+\sigma_x^2}+(1+\sigma_x^2)\frac{\mu^2}{(1+\sigma_x^2)^2}-(1+\sigma_x^2)\frac{\mu^2}{(1+\sigma_x^2)^2}+(1+\sigma_x^2)\frac{\mu^2}{1+\sigma_x^2}}{2\sigma_x^2}} dx \\
&=\frac{1}{2\pi\sigma_x}\biggintreals\gaussexp{\left(x-\frac{\mu_x}{1+\sigma_x^2}\right)^2}{\frac{\sigma_x^2}{1+\sigma_x^2}} \gaussexp{\mu_x^2}{1+\sigma_x^2}dx \\
&=\frac{1}{\sqrt{2\pi}\sigma_x}\sqrt{\frac{\sigma_x^2}{1+\sigma_x^2}}\gaussexp{\mu_x^2}{1+\sigma_x^2}\biggintreals \gaussfrac{\frac{\sigma_x^2}{1+\sigma_x^2}}\gaussexp{\left(x-\frac{\mu_x}{1+\sigma_x^2}\right)^2}{\frac{\sigma_x^2}{1+\sigma_x^2}}dx\\
&=\frac{1}{\sqrt{2\pi(1+\sigma_x^2)}}\gaussexp{\mu_x^2}{1+\sigma_x^2}
\end{split}
\end{equation}

\begin{equation}
	\begin{split}
\meanof{\phi(x)x}&=\frac{1}{\sqrt{2\pi}\sigma_x}\sqrt{\frac{\sigma_x^2}{1+\sigma_x^2}}\gaussexp{\mu_x^2}{1+\sigma_x^2}\intreals x\gaussfrac{\frac{\sigma_x^2}{1+\sigma_x^2}}\gaussexp{\left(x-\frac{\mu_x}{1+\sigma_x^2}\right)^2}{\frac{\sigma_x^2}{1+\sigma_x^2}}dx\\
&=\frac{1}{\sqrt{2\pi(1+\sigma_x^2)}}\gaussexp{\mu_x^2}{1+\sigma_x^2}\cdot\frac{\mu_x}{1+\sigma_x^2}
\end{split}
\end{equation}

\begin{equation}
	\begin{split}
\meanof{\phi(x)y}&=\left(\mu_y-\frac{\sigma_{xy}}{\sigma^2_y}\mu_x\right)\frac{1}{\sqrt{2\pi(1+\sigma_x^2)}}\gaussexp{\mu_x^2}{1+\sigma_x^2}+\frac{\sigma_{xy}}{\sigma^2_y}\frac{1}{\sqrt{2\pi(1+\sigma_x^2)}}\gaussexp{\mu_x^2}{1+\sigma_x^2}\frac{\mu_x}{1+\sigma_x^2} \\
&=\frac{1}{\sqrt{2\pi(1+\sigma_x^2)}}\gaussexp{\mu_x^2}{1+\sigma_x^2}\cdot\left(\mu_y-\frac{\sigma_{xy}}{\sigma^2_y}\frac{\mu_x\sigma_x^2}{1+\sigma_x^2}\right)
\end{split}
\end{equation}
Plugging back in the distributional parameters from \eqref{univdists}, we get
\begin{equation}\label{probitsolution}
	\begin{split}
		&\meanofwrt{g'(f(\fatx_{(r)}))\cdot f^{\partial_k}(\fatx_{(r)})}{g'(f(\fatx_{(r)}))\cdot f^{\partial_k}(\fatx_{(r)})}\\
		&\quad\quad=\frac{1}{\sqrt{2\pi(1+\krfatfU)}}\gaussexp{\mrfatfU^2}{1+\krfatfU}\cdot\left(\mrpkpfatfU-\frackkppk\frac{\mrfatfU\krfatfU}{1+\krfatfU}\right)
	\end{split}
\end{equation}

By plugging \eqref{probitsolution} into \eqref{meanofigmulti}, we get the proclaimed result:
\begin{equation}
\begin{split}
	\meanofwrt{\hat{IG}(\fatx,\fxb)_k^{g\circ f}}{q(f(\fatx_{(1...R)}),f^{\partial_k}(\fatx_{(1...R)}))} &= \frac{\fatx_k-\fxb_k}{R}\sum_{r=1}^R \meanofwrt{g'(f(\fatx_{(r)}))\cdot f^{\partial_k}(\fatx_{(r)})}{q(f(\fatx_{(r)}),f^{\partial_k}(\fatx_{(r)}))} \\
&=\frac{\fatx_k-\fxb_k}{R}\sum_{r=1}^R\frac{1}{\sqrt{2\pi(1+\krfatfU)}}\gaussexp{\mrfatfU^2}{1+\krfatfU}\\
	&\quad\quad\quad\quad\cdot\left(\mrpkpfatfU-\frackkppk\frac{\mrfatfU\krfatfU}{1+\krfatfU}\right)
\end{split}
\end{equation}

\subsection{C.3 Derivation for the $\exp$ inverse link-function}
For natural exponential inverse link-function, 

\begin{equation}
g(f(\fatx_{(r)}))=\exp\left(f(\fatx_{(r)})\right),
\end{equation}

the partial derivative for given $\fatx_{(r)}$ is

\begin{equation}
\frac{\partial}{\partial\fatx_k}g(f(\fatx_{(r)}))=\exp\left(f(\fatx_{(r)})\right)\cdot f^{\partial_k}(\fatx_{(r)}).
\end{equation}

For generic Gaussian distribution, we have

\begin{equation}
	\begin{split}
\meanof{g'(x)y}&=\meanof{\expof{x}y} \\
&=\meanof{\expof{x}\left(\mu_y+\frac{\sigma_{xy}}{\sigma^2_y}\left(x-\mu_x\right)\right)} \\
&=\left(\mu_y-\frac{\sigma_{xy}}{\sigma^2_y}\mu_x\right)\meanof{\expof{x}}+\frac{\sigma_{xy}}{\sigma^2_y}\meanof{\expof{x}x}
	\end{split}
\end{equation}

\begin{equation}
	\begin{split}
\meanof{\expof{x}}&=\intreals \expof{x} \gaussdens{(x-\mu_x)^2}{\sigma_x^2} dx \\
&=\intreals \frac{1}{2\pi\sigma_x}\gaussexp{x^2-2x\mu_x+\mu_x^2-2\sigma_x^2x}{\sigma_x^2} dx \\
&=\intreals \frac{1}{2\pi\sigma_x}\gaussexp{x^2-2x(\mu_x+\sigma_x^2)+\mu_x^2}{\sigma_x^2} dx \\
&=\intreals \frac{1}{2\pi\sigma_x}\gaussexp{x^2-2x(\mu_x+\sigma_x^2)+(\mu_x+\sigma_x^2)^2-(\mu_x+\sigma_x^2)^2+\mu_x^2}{\sigma_x^2} dx \\
&=\intreals \frac{1}{2\pi\sigma_x}\gaussexp{\left(x-\left(\mu_x+\sigma_x^2\right)\right)^2-2\mu_x\sigma_x^2-\sigma_x^4}{\sigma_x^2} dx \\
&=\expof{\mu_x+0.5\sigma_x^2}\intreals \frac{1}{2\pi\sigma_x}\gaussexp{\left(x-\left(\mu_x+\sigma_x^2\right)\right)^2}{\sigma_x^2} dx \\
&=\expof{\mu_x+0.5\sigma_x^2}dx
\end{split}
\end{equation}

\begin{equation}
	\begin{split}
\meanof{\expof{x}x}&=\expof{\mu_x+0.5\sigma_x^2}\intreals x \frac{1}{2\pi\sigma_x}\gaussexp{\left(x-\left(\mu_x+\sigma_x^2\right)\right)^2}{\sigma_x^2}\\
	&=\expof{\mu_x+0.5\sigma_x^2}\cdot\left(\mu_x+\sigma_x^2\right)
\end{split}
\end{equation}

\begin{equation}
	\begin{split}
\meanof{\expof{x}y}&=\left(\mu_y-\frac{\sigma_{xy}}{\sigma^2_y}\mu_x\right)\expof{\mu_x+0.5\sigma_x^2}+\frac{\sigma_{xy}}{\sigma^2_y}\expof{\mu_x+0.5\sigma_x^2}\cdot\left(\mu_x+\sigma_x^2\right) \\
&=\expof{\mu_x+0.5\sigma_x^2}\cdot\left(\mu_y+\frac{\sigma_{xy}}{\sigma^2_y}\sigma_x^2\right)
\end{split}
\end{equation}

Plugging back in the distributional parameters from \eqref{univdists}, we get

\begin{equation}\label{expsolution}
\meanofwrt{g'(f(\fatx_{(r)}))\cdot f^{\partial_k}(\fatx_{(r)})}{q(f\fatx_{(r)},f^{\partial_k}(\fatx_{(r)}))}=\expof{\mrfatfU+0.5\krfatfU}\cdot\left(\mrpkpfatfU+\frackkppk\krfatfU\right)
\end{equation}

By plugging \eqref{expsolution} into \eqref{meanofigmulti}, we get the proclaimed result:
\begin{equation}
\begin{split}
	\meanofwrt{\hat{IG}(\fatx,\fxb)_k^{g\circ f}}{q(f(\fatx_{(1...R)}),f^{\partial_k}(\fatx_{(1...R)}))} &= \frac{\fatx_k-\fxb_k}{R}\sum_{r=1}^R \meanofwrt{g'(f(\fatx_{(r)}))\cdot f^{\partial_k}(\fatx_{(r)})}{q(f(\fatx_{(r)}),f^{\partial_k}(\fatx_{(r)}))} \\
&=\frac{\fatx_k-\fxb_k}{R}\sum_{r=1}^R\expof{\mrfatfU+0.5\krfatfU}\cdot\left(\mrpkpfatfU+\frackkppk\krfatfU\right)
\end{split}
\end{equation}

\subsection{C.4 Derivation for the Sigmoid/Logistic inverse link-function}
The results in this and the next subsection will make use of the following Lemma:
\begin{lemma}
Let $\fatx\in\mathbb{R}^N$ denote a random vector with mean vector $\fatmu_\fatx$, $f,g:\mathbb{R}^N\mapsto\mathbb{R}$, $f$ and $g$ at least twice differentiable,and $h=f+g$. Then, the second-order Taylor approximation around the mean of $h(\fatx)$, 
\begin{equation}
\meanof{h(\fatx)}\approx T_2\meanof{h(\fatx)}=h(\fatmu_\fatx)+0.5\meanof{\fatx-\fatmu_\fatx}^T\nabla^2 h(\fatmu_\fatx)\meanof{\fatx-\fatmu_\fatx},
\end{equation}
is equivalent to the sum of second order Taylor approximations
\begin{equation}
T_2\meanof{h(\fatx)}=T_2\meanof{f(\fatx)}+T_2\meanof{g(\fatx)}
\end{equation}
\end{lemma}

\begin{proof}
	According to \cite{konigsberger2006analysis}, for any $C^p$ function, we have around an expansion point $\fata$:
	\begin{equation}
		\begin{split}
			& h(\fatx)=T_p h(\fatx; \fata)+o(||\fatx-\fata||^p) \\
			&= \sum_{k=1}^p \frac{1}{k!}d^{(k)}h(\fata)(\fatx-\fata)^k+o(||\fatx-\fata||^p) \\
			&= \sum_{k=1}^p \frac{1}{k!}d^{(k)}(f(\fata)+g(\fata))(\fatx-\fata)^k+o(||\fatx-\fata||^p) \\
			&= \sum_{k=1}^p \frac{1}{k!}d^{(k)}f(\fata)(\fatx-\fata)^k+\sum_{k=1}^p \frac{1}{k!}d^{(k)}g(\fata)(\fatx-\fata)^k+o(||\fatx-\fata||^p) \\
			&= T_p f(\fatx; \fata)+T_p g(\fatx; \fata)+o(||\fatx-\fata||^p),
		\end{split}
	\end{equation}
	where the properties of partial derivatives allow us to exchange summation and the $k$-th total differential, $d^{(k)}$.

Therefore, we have shown that a Taylor series approximation of any order for a sum of functions is equivalent to the sum of Taylor approximations of the individual functions. Then, expanding $h$ around $\fatmu_\fatx$, applying the above result and the expectation operation concludes the proof.
\end{proof}

We now proceed with deriving a Taylor approximation for the Sigmoid inverse link-function:
For sigmoid inverse link-function, 

\begin{equation}
g(f(\fatx_{(r)}))=s\left(f(\fatx_{(r)})\right),
\end{equation}

the partial derivative for given $\fatx_{(r)}$ is

\begin{equation}
\frac{\partial}{\partial\fatx_k}g(f(\fatx_{(r)}))=s'\left(f(\fatx_{(r)})\right)\cdot f^{\partial_k}(\fatx_{(r)}).
\end{equation}

For generic Gaussian distribution, we have

\begin{equation}\label{meanofsigmoidforappr}
	\begin{split}
\meanof{g'(x)y}&=\meanof{s'(x)y} \\
&=\meanof{s'(x)\left(\mu_y+\frac{\sigma_{xy}}{\sigma^2_y}\left(x-\mu_x\right)\right)}
		\end{split}
\end{equation}

To approximate \eqref{meanofsigmoidforappr} via second-order Taylor approximation, we need the second derivative of

\begin{equation}\label{originalfun}
f(x) = s'(x)\left(\mu_y+\frac{\sigma_{xy}}{\sigma^2_y}\left(x-\mu_x\right)\right)
\end{equation}

which is

\begin{equation}
f'(x) = s''(x)\mu_y + \frac{\sigma_{xy}}{\sigma^2_y}s''(x)x + \frac{\sigma_{xy}}{\sigma^2_y}s'(x) - \frac{\sigma_{xy}}{\sigma^2_y}s''(x)\mu_x
\end{equation}

\begin{equation}\label{secondorderfun}
f''(x)=s'''(x)\mu_y + \frac{\sigma_{xy}}{\sigma^2_y}s'''(x)x + \frac{\sigma_{xy}}{\sigma^2_y}s''(x) + \frac{\sigma_{xy}}{\sigma^2_y}s''(x) - \frac{\sigma_{xy}}{\sigma^2_y}s'''(x)\mu_x.
\end{equation}

Next, we plug in $\mu_x$ into \eqref{originalfun},

\begin{equation}
f(\mu_x)=s'(\mu_x)\mu_y,
\end{equation}

and \eqref{secondorderfun},

\begin{equation}
f''(\mu_x)=s'''(\mu_x)\mu_y + 2\frac{\sigma_{xy}}{\sigma^2_y}s''(\mu_x)
\end{equation}

to obtain the approximation

\begin{equation}
\meanof{s'(x)y}\approx s'(\mu_x)\mu_y + \frac{1}{2}s'''(\mu_x)\mu_y + \frac{\sigma_{xy}}{\sigma^2_y}s''(\mu_x).
\end{equation}

Plugging back in the distributional parameters from \eqref{univdists}, we get

\begin{equation}\label{sigmoidsolution}
s'(\mrfatfU)\mrpkpfatfU + \frac{1}{2}s'''(\mrfatfU)\mrpkpfatfU + \frackkppk s''(\mrfatfU)
\end{equation}

By plugging \eqref{sigmoidsolution} into \eqref{meanofigmulti}, we get the proclaimed result:
\begin{equation}
\begin{split}
	\meanofwrt{\hat{IG}(\fatx,\fxb)_k^{g\circ f}}{q(f(\fatx_{(1...R)}),f^{\partial_k}(\fatx_{(1...R)}))} &= \frac{\fatx_k-\fxb_k}{R}\sum_{r=1}^R \meanofwrt{g'(f(\fatx_{(r)}))\cdot f^{\partial_k}(\fatx_{(r)})}{q(f(\fatx_{(r)}),f^{\partial_k}(\fatx_{(r)}))} \\
	&\approx\frac{\fatx_k-\fxb_k}{R}\sum_{r=1}^R\left(s'(\mrfatfU)\mrpkpfatfU + \frac{1}{2}s'''(\mrfatfU)\mrpkpfatfU + \vphantom{\frackkppk s''(\mrfatfU)}\right. \\
	&\quad\quad\quad\quad\quad\quad\quad\quad\quad\quad\left. + \frackkppk s''(\mrfatfU)\right)
\end{split}
\end{equation}

\subsection{C.5 Derivation for the Softmax inverse link-function}
We begin the proof by recalling the gradient vector and hessian matrix for the softmax function with respect to $c$-th output and deterministic input $\fatx\in\mathbb{R}^C$:

\begin{equation}
	\left(\nabla S_c(\fatx)\right)_{(i)}=\begin{cases}
		S_c(\fatx)(1-S_c(\fatx)) & \text{if } i = c \\
		-S_c(\fatx)S_i(\fatx) & \text{if } i\neq c
	\end{cases}
\end{equation}

and

\begin{equation}
\begin{split}
	\left(\nabla^2 S_c(\fatx)\right)_{(ij)}=\begin{cases}
		S_c(\fatx)\cdot(1-3S_c(\fatx)+2S_c(\fatx)^2)& \text{if } i = j = c \\
		S_c(\fatx)S_i(\fatx)(2S_i(\fatx)-1) & \text{if } i = j \neq c \\
		S_j(\fatx)S_c(\fatx)(2S_c(\fatx)-1) & \text{if } i \neq j, i = c \\
		S_i(\fatx)S_c(\fatx)(2S_c(\fatx)-1) & \text{if } i \neq j, j = c \\
		2S_c(\fatx)S_i(\fatx)S_j(\fatx) & \text{if } i \neq j, i \neq c, j\neq c \\
	\end{cases}
\end{split}
\end{equation}

From \eqref{multiggptarget}, it follows that we need to approximate

\begin{equation}\label{softmaxtargetfun}
\begin{split}
	\sum_{j=1}^C\meanofwrt{\frac{\partial}{\partial f^{(j)}}S_c(\fatF(\fatx_{(r)})) \cdot f^{(j)\partial_k}(\fatx_{(r)})}{q(\fatF(\fatx_{(r)}))}&=\meanofwrt{\Si{c}(\fatF(\fatx_{(r)}))\left(1-\Si{c}(\fatF(\fatx_{(r)}))\right)\fipk{c}(\fatx_{(r)})}{q(\fatF(\fatx_{(r)}))} \\
	&- \sum_{j\in\{1,...,C\}\setminus c}\meanofwrt{\Si{c}(\fatF(\fatx_{(r)}))\Si{j}(\fatF(\fatx_{(r)}))\fipk{j}(\fatx_{(r)})}{q(\fatF(\fatx_{(r)}))}.
\end{split}
\end{equation}

According to Lemma 1, we have 

\begin{equation}
\begin{split}
	q\left(\begin{bmatrix}\fatF(\fatx_{(r)}) \\ f^{(j)\partial_k}(\fatx_{(r)})\end{bmatrix}\right) &= \mathcal{N}\left(\begin{bmatrix}\fatF(\fatx_{(r)}) \\ f^{(j)\partial_k}(\fatx_{(r)})\end{bmatrix}\Bigg| \begin{bmatrix}\fatm_{(r)|\fatf_U} \\ m^{(j)\partial_k}_{(r)|\fatf_U}\end{bmatrix},\begin{bmatrix}
		\fatK_{(r)|\fatf_U} & \fatk^{(j)\partial_k}_{(r)|\fatf_U} \\
		\left(\fatk^{(j)\partial_k}_{(r)|\fatf_U}\right)^T & k^{(j)\partial_k\partial_k'}_{(r)|\fatf_U} 
\end{bmatrix}\right)\\
&=\mathcal{N}\left(\begin{bmatrix}
		f_{\vphantom{(r)|\fatf_U}}^{(1)}(\fatx_{(r)})\\
\vdots \\
	f_{\vphantom{(r)|\fatf_U}}^{(j)}(\fatx_{(r)}) \\
\vdots \\
	f_{\vphantom{(r)|\fatf_U}}^{(C)}(\fatx_{(r)}) \\
	f_{\vphantom{(r)|\fatf_U}}^{(j)\partial_k}(\fatx_{(r)})
\end{bmatrix}\Bigg|\begin{bmatrix}
m^{(1)}_{(r)|\fatf_U} \\ 
\vdots \\ 
m^{(j)}_{(r)|\fatf_U} \\
\vdots \\
m^{(C)}_{(r)|\fatf_U} \\ 
m^{(j)\partial_k}_{(r)|\fatf_U}
\end{bmatrix}, \begin{bmatrix}
	k^{(1)}_{(r)|\fatf_U}  & \cdots & 0 & \cdots & 0 & 0 \\
	\vdots  & \ddots & \ddots & \ddots & \ddots & \vdots \\
	0 & \ddots & k^{(j)}_{(r)|\fatf_U} & \ddots & 0 & k^{(j)\partial_k}_{(r)|\fatf_U} \\
	\vdots  & \ddots & \ddots & \ddots & \ddots & \vdots \\
	0 & \cdots & 0 & \cdots & k^{(C)}_{(r)|\fatf_U} & 0 \\
	0 & \cdots &k^{(j)\partial_k}_{(r)|\fatf_U} & \cdots & 0 & k^{(j)\partial_k\partial_k'}_{(r)|\fatf_U} 
\end{bmatrix}\right)
\end{split}
\end{equation}

To get a more convenient notation for the following steps, we substitute $\begin{bmatrix}\fatF(\fatx_{(r)}) \\ f^{(j)\partial_k}(\fatx_{(r)})\end{bmatrix}$ with $\begin{bmatrix}\fatx \\ y\end{bmatrix}$ with corresponding distribution:

\begin{equation}
	\begin{split}
	q\left(\begin{bmatrix}\fatx \\ y\end{bmatrix}\right) &= \mathcal{N}\left(\begin{bmatrix}\fatx \\ y\end{bmatrix}\Bigg| \begin{bmatrix}\fmufx \\ \mu_y\end{bmatrix},\begin{bmatrix}
		\fatSigma_{\fatx\fatx} & \fatsigma_{\fatx y} \\
		\fatsigma_{\fatx y}^T & \sigma_{\fatx y} \\
	\end{bmatrix}\right)\\
	&\mathrel{\widehat{=}} q\left(\begin{bmatrix}\fatF(\fatx_{(r)})\\ f^{(j)\partial_k}(\fatx_{(r)})\end{bmatrix}\right)
\end{split}
\end{equation}

Notice in particular that $\fatSigma_{\fatx\fatx}$ is a diagonal matrix. We now split the problem of (Tayor-) approximating \eqref{softmaxtargetfun} into approximations of the individual sums which, according to Lemma 3, is equivalent to approximating the full sum at once.

For the lhs, $\meanofwrt{\Si{c}(\fatF(\fatx_{(r)}))\left(1-\Si{c}(\fatF(\fatx_{(r)}))\right)\fipk{c}(\fatx_{(r)})}{q(\fatF(\fatx_{(r)}))}$, we start by substituting in the simplified notation from above which yields

\begin{equation}
	\meanofwrt{\Si{c}(\fatx)\left(1-\Si{c}(\fatx)\right)\cdot y}{q(\fatx,y)}
\end{equation}

Applying Lemma 2, this simplifies to

\begin{equation}\label{lhssoftmaxtarget}
\begin{split}
\meanofwrt{\Si{c}(\fatx)\left(1-\Si{c}(\fatx)\right)\cdot y}{q(\fatx,y)}&=\meanofwrt{\Si{c}(\fatx)\left(1-\Si{c}(\fatx)\right)\cdot\meanof{y|\fatx}}{q(\fatx,y)} \\
&=\mathbb{E}_{q(\fatx,y)}[(\Si{c}(\fatx)-\Si{c}(\fatx)^2)\cdot(\mu_{y}+\underbrace{\left(\fatsigma_{\fatx y}\right)^T\fatSigma_{\fatx\fatx}^{-1}}_{:=\fatSigma}(\fatx-\fmufx))] \\
	&=\mathbb{E}_{q(\fatx,y)}[\underbrace{\Si{c}(\fatx)\mu_y - \Si{c}(\fatx)^2\mu_y + \Si{c}(\fatx)\fatSigma\fatx - \Si{c}(\fatx)^2\fatx - \Si{c}(\fatx)\fatSigma\fmufx + \Si{c}(\fatx)^2\fmufx}_{:=\tilde{H}_c(\fatx)}].
\end{split}
\end{equation}

To derive the second-order Taylor approximation for \eqref{lhssoftmaxtarget}, we need to treat the interior of the mean as a function $\tilde{H}_c$ whose input is the random vector $\fatx$. Later, we will re-substitute the original distributional parameters from $q(\fatF(\fatx_{(r)}), f^{(j)\partial_k}(\fatx_{(r)}))$ and denote the corresponding function as $H_{c(r)}^k$.

For now, recall the formula for the second-order Taylor approximation for function with multivariate input:

\begin{equation}
	T_2\meanof{\tilde{H}_c(\fatx)}=\tilde{H}_c(\mu_\fatx)+0.5\meanof{\left(\fatx-\mu_\fatx\right)^T\nabla^2 \tilde{H}_c(\mu_\fatx)\left(\fatx-\mu_\fatx\right)}
\end{equation}

Due to $\fatSigma_{\fatx\fatx}$ being diagonal, this simplifies to 

\begin{equation}
	T_2\meanof{\tilde{H}_c(\fatx)}=\tilde{H}_c(\mu_\fatx)+0.5\diag\left(\fatSigma_{\fatx\fatx}\right)^T\diag\left(\nabla^2 \tilde{H}_c(\mu_\fatx)\right)
\end{equation}

We continue with the gradient of $\tilde{H}_c(\fatx)$:

\begin{equation}
\begin{split}
\nabla \tilde{H}_c(\fatx) &= \nabla S_c(\fatx)\mu_y - S_c(\fatx)\nabla S_c(\fatx)^2\mu_y \\
&\quad\quad + (\fatSigma \fatx)\nabla S_c(\fatx) + \fatSigma^T S_c(\fatx) - (\fatSigma\fatx)^2 S_c(\fatx)\nabla S_c(\fatx) - \fatSigma^T S_c(\fatx)^2 \\
&\quad\quad - (\fatSigma\fmufx)\nabla S_c(\fatx) + (\fatSigma \fmufx)2S_c(\fatx)\nabla S_c(\fatx)
\end{split}
\end{equation}

The Hessian is

\begin{equation}
\begin{split}
\nabla^2 \tilde{H}_c(\fatx) &= \nabla^2 S_c(\fatx)\mu_y - \nabla S_c(\fatx)\nabla S_c(\fatx)^T 2\mu_y - S_c(\fatx)\nabla^2 S_c(\fatx) 2\mu_y \\
&\quad\quad + \fatSigma^T \nabla S_c(\fatx)^T + \nabla^2 S_c(\fatx) (\fatSigma\fatx) + \nabla S_c(\fatx)\fatSigma - 2 S_c(\fatx)\fatSigma^T \nabla S_c(\fatx)^T  \\
&\quad\quad - 2 S_c(\fatx)\nabla S_c(\fatx)\fatSigma - 2(\fatSigma\fatx)\nabla S_c(\fatx)\nabla S_c(\fatx)^T - 2(\fatSigma\fatx) S_c(\fatx)\nabla^2 S_c(\fatx) \\
&\quad\quad - \nabla^2 S_c(\fatx)(\fatSigma\fmufx) + 2 (\fatSigma\fmufx) \nabla S_c(\fatx)\nabla S_c(\fatx)^T + 2(\fatSigma \fmufx)S_c(\fatx) \nabla^2 S_c(\fatx).
\end{split}
\end{equation}

Then, we plug in $\fmufx$ and simplify:

\begin{equation}
\begin{split}
\nabla^2 \tilde{H}_c(\fmufx) &= \nabla^2 S_c(\fmufx)\mu_y - \nabla S_c(\fmufx)\nabla S_c(\fmufx)^T 2\mu_y - S_c(\fmufx)\nabla^2 S_c(\fmufx) 2\mu_y \\
&\quad\quad + \fatSigma^T \nabla S_c(\fmufx)^T + \cancel{\nabla^2 S_c(\fmufx) (\fatSigma\fmufx)} + \nabla S_c(\fmufx)\fatSigma - 2 S_c(\fmufx)\fatSigma^T \nabla S_c(\fmufx)^T  \\
&\quad\quad - 2 S_c(\fmufx)\nabla S_c(\fmufx)\fatSigma - \cancel{2(\fatSigma\fmufx)\nabla S_c(\fmufx)\nabla S_c(\fmufx)^T} - \cancel{2(\fatSigma\fmufx) S_c(\fmufx)\nabla^2 S_c(\fmufx)} \\
&\quad\quad - \cancel{\nabla^2 S_c(\fmufx)(\fatSigma\fmufx)} + \cancel{2 (\fatSigma\fmufx) \nabla S_c(\fmufx)\nabla S_c(\fmufx)^T} + \cancel{2(\fatSigma \fmufx)S_c(\fmufx) \nabla^2 S_c(\fmufx)} \\
& \\
&= \nabla^2 S_c(\fmufx)\mu_y - \nabla S_c(\fmufx)\nabla S_c(\fmufx)^T 2\mu_y - S_c(\fmufx)\nabla^2 S_c(\fmufx) 2\mu_y \\
&\quad\quad + \fatSigma^T \nabla S_c(\fmufx)^T + \nabla S_c(\fmufx)\fatSigma - 2 S_c(\fmufx)\fatSigma^T \nabla S_c(\fmufx)^T - S_c(\fmufx)\nabla S_c(\fmufx)\fatSigma \\
\end{split}
\end{equation}

Resubstituting to obtain $H_{c(r)}^k$ and evaluating at $\fatm_{(r)|\fatf_U}$, we get

\begin{equation}\label{hcfun}
	\begin{split}
		H_{c(r)}^k(\fatm_{(r)|\fatf_U})&=\Si{c}(\fatm_{(r)|\fatf_U})m^{(j)\partial_k}_{(r)|\fatf_U} - \Si{c}(\fatm_{(r)|\fatf_U})^2m^{(j)\partial_k}_{(r)|\fatf_U} + \Si{c}(\fatm_{(r)|\fatf_U})\left(\fatk^{(j)\partial_k}_{(r)|\fatf_U}\right)^T\fatK_{(r)|\fatf_U}^{-1}\fatm_{(r)|\fatf_U} \\
		&\quad\quad- \Si{c}(\fatm_{(r)|\fatf_U})^2\fatm_{(r)|\fatf_U} - \Si{c}(\fatm_{(r)|\fatf_U})\left(\fatk^{(j)\partial_k}_{(r)|\fatf_U}\right)^T\fatK_{(r)|\fatf_U}^{-1}\fatm_{(r)|\fatf_U} + \Si{c}(\fatm_{(r)|\fatf_U})^2\fatm_{(r)|\fatf_U}
	\end{split}
\end{equation}

\begin{equation}\label{nablahcfun}
\begin{split}
\nabla^2 H_{c(r)}^k(\fatm_{(r)|\fatf_U}) &= \nabla^2 S_c(\fatm_{(r)|\fatf_U})m^{(j)\partial_k}_{(r)|\fatf_U} - \nabla S_c(\fatm_{(r)|\fatf_U})\nabla S_c(\fatm_{(r)|\fatf_U})^T 2m^{(j)\partial_k}_{(r)|\fatf_U} \\
	&\quad\quad - S_c(\fatm_{(r)|\fatf_U})\nabla^2 S_c(\fatm_{(r)|\fatf_U}) 2m^{(j)\partial_k}_{(r)|\fatf_U} + \left(\left(\fatk^{(j)\partial_k}_{(r)|\fatf_U}\right)^T\fatK_{(r)|\fatf_U}^{-1}\right)^T \nabla S_c(\fatm_{(r)|\fatf_U})^T \\
	&\quad\quad + \nabla S_c(\fatm_{(r)|\fatf_U})\left(\fatk^{(j)\partial_k}_{(r)|\fatf_U}\right)^T\fatK_{(r)|\fatf_U}^{-1}  \\
	&\quad\quad- 2 S_c(\fatm_{(r)|\fatf_U})\left(\left(\fatk^{(j)\partial_k}_{(r)|\fatf_U}\right)^T\fatK_{(r)|\fatf_U}^{-1}\right)^T \nabla S_c(\fatm_{(r)|\fatf_U})^T \\
	&\quad\quad - S_c(\fatm_{(r)|\fatf_U})\nabla S_c(\fatm_{(r)|\fatf_U})\left(\fatk^{(j)\partial_k}_{(r)|\fatf_U}\right)^T\fatK_{(r)|\fatf_U}^{-1} \\
\end{split}
\end{equation}

The approximation of the lhs of \eqref{softmaxtargetfun} is therefore
\begin{equation}
	\begin{split}
	&\meanofwrt{\Si{c}(\fatF(\fatx_{(r)}))\left(1-\Si{c}(\fatF(\fatx_{(r)}))\right)f^{(c)\partial_k}(\fatx_{(r)})}{q(\fatF(\fatx_{(r)}))} \\
	&\quad\quad\approx H_{c(r)}^k(\fatm_{(r)|\fatf_U})+0.5\diag(\fatK_{(r)|\fatf_U})^T\diag(\nabla^2 H_{c(r)}^k(\fatm_{(r)|\fatf_U}))
	\end{split}
\end{equation}

with $H_{c(r)}^k(\fatm_{(r)|\fatf_U}), \nabla^2 H_{c(r)}^k(\fatm_{(r)|\fatf_U})$ defined as in \eqref{hcfun}, \eqref{nablahcfun}.

To approximate the rhs of \eqref{softmaxtargetfun}, we proceed similarly to the lhs by approximating each summand mean individually. We begin with the substituted derivation of the corresponding Hessian matrix:

\begin{equation}
\begin{split}
\meanof{S_c(\fatx)S_j(\fatx)\cdot y}&=\meanof{S_c(\fatx)S_j(\fatx)\cdot \meanof{y|\fatx}} \\
&=\mathbb{E}[\underbrace{S_c(\fatx)S_j(\fatx)\mu_y+S_c(\fatx)S_j(\fatx)\fatSigma\fatx -S_c(\fatx)S_j(\fatx)\fatSigma\fmufx}_{:=\tilde{H}_{cj}(\fatx)}]
\end{split}
\end{equation}

Gradient and Hessian are then derived as follows:

\begin{equation}
\begin{split}
\nabla \tilde{H}_{cj}(\fatx) &= \nabla S_c(\fatx)S_j(\fatx)\mu_y + S_c(\fatx)\nabla S_j(x)\mu_y \\
&=\quad\quad + \nabla S_c(\fatx)S_j(\fatx)(\fatSigma\fatx) + S_c(\fatx)\nabla S_j(\fatx)(\fatSigma\fatx) + S_c(\fatx)S_j(\fatx)\fatSigma^T \\
&=\quad\quad - \nabla S_c(\fatx)S_j(\fatx)(\fatSigma\fmufx) - S_c(\fatx)\nabla S_j(\fatx)(\fatSigma\fmufx)
\end{split}
\end{equation}

\begin{equation}
\begin{split}
\nabla^2 \tilde{H}_{cj}(\fatx) &= \nabla S_c(\fatx)\nabla S_j(\fatx)^T \mu_y + \nabla^2 S_c(\fatx)S_j(\fatx)\mu_y + \nabla S_j(\fatx)\nabla S_c(\fatx)^T \mu_y + S_c(\fatx)\nabla^2 S_j(\fatx)\mu_y \\
&\quad\quad + \nabla^2 S_c(\fatx)S_j(\fatx)(\fatSigma\fatx) + \nabla S_c(\fatx)\nabla S_j(\fatx)^T (\fatSigma\fatx) + S_j(\fatx)(\nabla S_c(\fatx) \fatSigma) \\
&\quad\quad + S_c(\fatx)\nabla^2 S_j(\fatx)(\fatSigma\fatx) + \nabla S_j(\fatx)\nabla S_c(\fatx)^T (\fatSigma\fatx) + S_c(\fatx)(\nabla S_j(\fatx) \fatSigma) \\
&\quad\quad + S_j(\fatx) (\fatSigma^T\nabla S_c(\fatx)^T) + S_c(\fatx) (\fatSigma^T\nabla S_j(\fatx)^T)  \\
&\quad\quad - \nabla^2 S_c(\fatx)S_j(\fatx)(\fatSigma\fmufx) - \nabla S_c(\fatx)\nabla S_j(\fatx)^T (\fatSigma\fmufx) \\
&\quad\quad - \nabla S_j(\fatx)\nabla S_c(\fatx)^T (\fatSigma\fmufx) - S_c(\fatx)\nabla^2 S_j(\fatx)(\fatSigma\fmufx)
\end{split}
\end{equation}

Again, plugging in $\fmufx$ and simplifying:

\begin{equation}
\begin{split}
\nabla^2 \tilde{H}_{cj}(\fmufx) &= \nabla S_c(\fmufx)\nabla S_j(\fmufx)^T \mu_y + \nabla^2 S_c(\fmufx)S_j(\fmufx)\mu_y + \nabla S_j(\fmufx)\nabla S_c(\fmufx)^T \mu_y + S_c(\fmufx)\nabla^2 S_j(\fmufx)\mu_y \\
&\quad\quad + \cancel{\nabla^2 S_c(\fmufx)S_j(\fmufx)(\fatSigma\fmufx)} + \cancel{\nabla S_c(\fmufx)\nabla S_j(\fmufx)^T (\fatSigma\fmufx)} + S_j(\fmufx)(\nabla S_c(\fmufx) \fatSigma) \\
&\quad\quad + \cancel{S_c(\fmufx)\nabla^2 S_j(\fmufx)(\fatSigma\fmufx)} + \cancel{\nabla S_j(\fmufx)\nabla S_c(\fmufx)^T (\fatSigma\fmufx)} + S_c(\fmufx)(\nabla S_j(\fmufx) \fatSigma) \\
&\quad\quad + S_j(\fmufx) (\fatSigma^T\nabla S_c(\fmufx)^T) + S_c(\fmufx) (\fatSigma^T\nabla S_j(\fmufx))  \\
&\quad\quad - \cancel{\nabla^2 S_c(\fmufx)S_j(\fmufx)(\fatSigma\fmufx)} - \cancel{\nabla S_c(\fmufx)\nabla S_j(\fmufx)^T (\fatSigma\fmufx)} \\
&\quad\quad - \cancel{\nabla S_j(\fmufx)\nabla S_c(\fmufx)^T (\fatSigma\fmufx)} - \cancel{S_c(\fmufx)\nabla^2 S_j(\fmufx)(\fatSigma\fmufx)} \\
& \\
&= \nabla S_c(\fmufx)\nabla S_j(\fmufx)^T \mu_y + \nabla^2 S_c(\fmufx)S_j(\fmufx)\mu_y + \nabla S_j(\fmufx)\nabla S_c(\fmufx)^T \mu_y \\
&\quad\quad  + S_c(\fmufx)\nabla^2 S_j(\fmufx)\mu_y + S_j(\fmufx)(\nabla S_c(\fmufx) \fatSigma) + S_c(\fmufx)(\nabla S_j(\fmufx) \fatSigma) \\
&\quad\quad + S_j(\fmufx) (\fatSigma^T\nabla S_c(\fmufx)^T) + S_c(\fmufx) (\fatSigma^T\nabla S_j(\fmufx)) 
\end{split}
\end{equation}

\begin{equation}\label{hcjfun}
	\begin{split}
		H^k_{cj(r)}(\fatm_{(r)|\fatf_U}) &= S_c(\fatm_{(r)|\fatf_U})S_j(\fatm_{(r)|\fatf_U})m^{(j)\partial_k}_{(r)|\fatf_U}+S_c(\fatm_{(r)|\fatf_U})S_j(\fatm_{(r)|\fatf_U})\left(\fatk^{(j)\partial_k}_{(r)|\fatf_U}\right)^T\fatK_{(r)|\fatf_U}^{-1}\fatm_{(r)|\fatf_U} \\
		&\quad\quad-S_c(\fatm_{(r)|\fatf_U})S_j(\fatm_{(r)|\fatf_U})\left(\fatk^{(j)\partial_k}_{(r)|\fatf_U}\right)^T\fatK_{(r)|\fatf_U}^{-1}\fatm_{(r)|\fatf_U}
	\end{split}
\end{equation}

\begin{equation}\label{nablahcjfun}
\begin{split}
	\nabla^2 H^k_{cj(r)}(\fatm_{(r)|\fatf_U}) &= \nabla S_c(\fatm_{(r)|\fatf_U})\nabla S_j(\fatm_{(r)|\fatf_U})^T m^{(j)\partial_k}_{(r)|\fatf_U} + \nabla^2 S_c(\fatm_{(r)|\fatf_U})S_j(\fatm_{(r)|\fatf_U})m^{(j)\partial_k}_{(r)|\fatf_U} \\
	&\quad\quad + \nabla S_j(\fatm_{(r)|\fatf_U})\nabla S_c(\fatm_{(r)|\fatf_U})^T m^{(j)\partial_k}_{(r)|\fatf_U} + S_c(\fatm_{(r)|\fatf_U})\nabla^2 S_j(\fatm_{(r)|\fatf_U})m^{(j)\partial_k}_{(r)|\fatf_U}\\
&\quad\quad   + S_j(\fatm_{(r)|\fatf_U})(\nabla S_c(\fatm_{(r)|\fatf_U}) \left(\fatk^{(j)\partial_k}_{(r)|\fatf_U}\right)^T\fatK_{(r)|\fatf_U}^{-1}) \\
	&\quad\quad + S_c(\fatm_{(r)|\fatf_U})(\nabla S_j(\fatm_{(r)|\fatf_U}) \left(\fatk^{(j)\partial_k}_{(r)|\fatf_U}\right)^T\fatK_{(r)|\fatf_U}^{-1}) \\
	&\quad\quad + S_j(\fatm_{(r)|\fatf_U}) \left(\left(\fatk^{(j)\partial_k}_{(r)|\fatf_U}\right)^T\fatK_{(r)|\fatf_U}^{-1}\right)^T\nabla S_c(\fatm_{(r)|\fatf_U})^T) \\
	&\quad\quad + S_c(\fatm_{(r)|\fatf_U}) \left(\left(\fatk^{(j)\partial_k}_{(r)|\fatf_U}\right)^T\fatK_{(r)|\fatf_U}^{-1}\right)^T\nabla S_j(\fatm_{(r)|\fatf_U})) 
\end{split}
\end{equation}

The approximation of the rhs of \eqref{softmaxtargetfun} is therefore
\begin{equation}
	\begin{gathered}
	\sum_{j\in\{1,...,C\}\setminus c}\meanofwrt{\Si{c}(\fatF(\fatx_{(r)}))\Si{j}(\fatF(\fatx_{(r)}))\fipk{j}(\fatx_{(r)})}{q(\fatF(\fatx_{(r)}))}\\
		\approx \sum_{j\in\{1,...,C\}\setminus c} H^k_{cj(r)}(\fatm_{(r)|\fatf_U})+0.5\diag(\fatK_{(r)|\fatf_U})^T\diag(\nabla^2 H^k_{cj(r)}(\fatm_{(r)|\fatf_U}))
	\end{gathered}
\end{equation}

with $H^k_{cj(r)}(\fatm_{(r)|\fatf_U}), \nabla^2 H^k_{cj(r)}(\fatm_{(r)|\fatf_U})$ defined as in \eqref{hcjfun}, \eqref{nablahcjfun}.

The full approximation is therefore

\begin{equation}
\begin{gathered}
	\sum_{j=1}^C\meanofwrt{\frac{\partial}{\partial f^{(j)}}S_c(\fatF(\fatx_{(r)})) \cdot f^{(j)\partial_k}(\fatx_{(r)})}{q(\fatF(\fatx_{(r)}))}\\
	\approx  H_{c(r)}^k(\fatm_{(r)|\fatf_U})+0.5\diag(\fatK_{(r)|\fatf_U})^T\diag(\nabla^2 H_{c(r)}^k(\fatm_{(r)|\fatf_U})) \\
	- \sum_{j\in\{1,...,C\}\setminus c} H_{cj(r)}^k(\fatm_{(r)|\fatf_U})+0.5\diag(\fatK_{(r)|\fatf_U})^T\diag(\nabla^2 H_{cj(r)}^k(\fatm_{(r)|\fatf_U}))
\end{gathered}
\end{equation}

and, finally, the approximation of the whole Integrated Gradient for the $k$-th feature attribution:

\begin{equation}
	\begin{gathered}
\meanofwrt{\hat{IG}(\fatx,\fatx')_k^{S_c\circ F}}{q(\fatF(\fatx^{(1)}),...,\fatF(\fatx^{(R)}))} \\
		\approx \frac{\fatx_k-\fatx_k'}{R}\left(\sum_{r=1}^R H_{c(r)}^k(\fatm_{(r)|\fatf_U})+0.5\diag(\fatK_{(r)|\fatf_U})^T\diag(\nabla^2 H_{c(r)}^k(\fatm_{(r)|\fatf_U}))\vphantom{\sum_{j\in\{1,...,C\}\setminus c}}\right. \\
		\left.- \sum_{j\in\{1,...,C\}\setminus c} H_{cj(r)}^k(\fatm_{(r)|\fatf_U})+0.5\diag(\fatK_{(r)|\fatf_U})^T\diag(\nabla^2 H_{cj(r)}^k(\fatm_{(r)|\fatf_U}))\right) \\
	\end{gathered}
\end{equation}

\end{proof}

\section{D. Proof of Theorem 2}
\begin{proof}[\unskip\nopunct]
The proof is straightforward but requires exchange of integration and expectation for which we need to apply the Fubini-Tonelli theorem and ensure absolute integrability of the target function. We begin with the proof itself and discuss the supporting Lemmas afterwards. First, let $\fatx_\alpha=\fatx_B-\alpha\cdot(\fatx-\fatx^B); \fatx,\fatx^B\in\mathbb{R}^M$. For the multi-output case, we have
	\begin{equation}\label{iggpresult}
	\begin{split}
		&\meanofwrt{G_c(F(\fatx))}{q(F(\fatx))}-\meanofwrt{G_c(F(\fatx^B))}{q(F(\fatx^B))} \\
		&=\meanofwrt{G_c(F(\fatx))-G_c(F(\fatx^B))}{q(F(\fatx),F(\fatx^B))} \\
		&=\meanofwrt{\sum_{k=1}^M\left(\fatx_{(k)}-\fatx^B_{(k)}\right)\int_0^1\nabla G_c(F(\fatx_\alpha))^T F^{\Jk}(\fatx_\alpha)d\alpha}{q(F(\fatx_\alpha),F^{\Jk}(\fatx_\alpha))} \\
		&=\sum_{k=1}^M\left(\fatx_{(k)}-\fatx^B_{(k)}\right)\meanofwrt{\int_0^1\nabla G_c(F(\fatx_\alpha))^T F^{\Jk}(\fatx_\alpha)d\alpha}{q(F(\fatx_\alpha),F^{\Jk}(\fatx_\alpha))} \\
		&=\sum_{k=1}^M\left(\fatx_{(k)}-\fatx^B_{(k)}\right)\int_0^1\meanofwrt{\nabla G_c(F(\fatx_\alpha))^T F^{\Jk}(\fatx_\alpha)}{q(F(\fatx_\alpha),F^{\Jk}(\fatx_\alpha))}d\alpha \\
		&=\sum_{k=1}^M\left(\fatx_{(k)}-\fatx^B_{(k)}\right)\lim_{R\rightarrow\infty}\frac{1}{R}\sum_{r=1}^R\meanofwrt{\nabla G_c(F(\fatx_r))^T F^{\Jk}(\fatx_r)}{q(F(\fatx_r),F^{\Jk}(\fatx_r))}d\alpha
	\end{split}
\end{equation}
	The single-output case follows the exact same steps. Notice the exchange of integration and expectation in the fourth equality. To prove absolute integrability and therefore validity of this step, we continue as follows: We start with Lemma 4, which provides a moment bound for the GP transformed by a polynomially bounded inverse link-function. Lemma 5 proves a moment bound for the supremum of the absolute of a Gaussian random vector. While both bounds could probably be improved further, we merely rely on them to prove absolute integrability for \eqref{iggpresult} in Lemma 6. 

\begin{lemma}
	Let $\fatx\in\mathbb{R}^M\sim\mathcal{N}(\fatmu,\fatSigma)$ denote a multivariate Gaussian vector, where $\fatSigma=\fatL\fatL^T$ and $G:\mathbb{R}^M\mapsto\mathbb{R}^C$ such that, for all $\faty\in\mathbb{R}^M$
	\begin{equation}
		\sup_{j\in 1,...,C}\left(|G(\faty)|\right)_{(j)}\leq D\left(1+\left(\sum_{j=1}^M|\faty_{(j)}|\right)^r\right),\quad D>0,r\geq 0.
	\end{equation}
	It follows that
	\begin{equation}
		\meanof{\left(\sup_{j\in 1,...,C} \left(|G(\fatx)|\right)_{(j)}\right)^2}\leq 2D^2+2D^2M^2r!(4P)^r
	\end{equation}
	where $P=\max_{j\in 1,...,M}\sqrt{\sum_{i=1}^M L_{(ij)}^2}$.
\end{lemma}
\begin{proof}
	Let $\tilde{\fatx}\sim\mathcal{N}(\fatzero,\fatI)$ and, for any $\tilde{\fatx},\tilde{\faty}\in\mathbb{R}^M$,
	\begin{equation}
		f(\tilde{\fatx})=\max_{j\in 1,...,M}\left(\absof{(L\tilde{\fatx}+\fatmu)}\right)_{(j)}.
	\end{equation}
	We have
	\begin{equation}
		f(\tilde{\fatx})\overset{d}{=}\max_{j\in 1,...,M}\left(\absof{\fatx}\right)_{(j)}
	\end{equation}
	and
	\begin{equation}
		\begin{split}
			&\absof{f(\tilde{\fatx})-f(\tilde{\faty})} \\
			&=\absof{\normof{L\tilde{\fatx}+\fatmu}{\infty}-\normof{L\tilde{\faty}+\fatmu}{\infty}}\\
			&\leq \max_{j\in 1,...,M}\sqrt{\sum_{i=1}^C L_{(ij)}^2}\normof{\tilde{\fatx}-\tilde{\faty}}{2}
		\end{split}
	\end{equation}
	Hence, $\max_{j\in 1,...,M}\absof{(L\tilde{\fatx}+\fatmu)_{(j)}}$ is Lipschitz with constant $P=\max_{j\in 1,...,M}\sqrt{\sum_{i=1}^C L_{(ij)}^2}$. According to \cite{boucheron}, Theorem 5.5 and Theorem 2.1, this implies
\begin{equation}
	\meanof{\left(\max_{j\in 1,...,M}\left(\absof{(\fatx)}\right)_{(j)}\right)^{2r}}=\meanof{\left(\max_{j\in 1,...,M}\left(\absof{(L\tilde{\fatx}+\fatmu)}\right)_{(j)}\right)^{2r}}\leq r!(4P)^r
\end{equation}
	Using this property, we can derive the desired bound as follows:
	\begin{equation}
		\begin{split}
			&\meanof{\left(\sup_{j\in 1,...,M} \left(|G(\fatx)|\right)_{(j)}\right)^2} \\
			&\leq\meanof{\left(D+D\left(\sum_{j=1}^M\absof{\fatx_{(j)}}\right)^r\right)^2} \\
			&\leq\meanof{\left(D+DC\left(\max_{j\in 1,...,M}\left(\absof{\fatx}\right)_{(j)}\right)^r\right)^2} \\
			&\leq 2D^2+2D^2M^2\meanof{\left(\max_{j\in 1,...,M}\left(\absof{L\tilde{\fatx}+\fatmu}\right)_{(j)}\right)^{2r}} \\
		&\leq 2D^2+2D^2M^2r!(4P)^r
		\end{split}
	\end{equation}
\end{proof}

\begin{lemma}
	Let $f(t),t\in T$ denote a GP over an arbitrary, countably finite index set $T$, with mean function $m(\cdot)$ where $\sigma^2_T\hat{=}\sup_{t\in T}\meanof{f(t)^2}<\infty$, $M\hat{=}\sup_{t\in T}|m(t)|=M<\infty$. Let, additionally, $\upsilon\hat{=}\min_{t\in T}\Var\left(f(t)\right)$. It then follows that

\begin{equation}
	\meanof{\left(\sup_{t\in T}|f(t)|\right)^2} < \infty,
\end{equation}
\end{lemma}
\begin{proof}
Let $f(t)=\tilde{f}(t)+m(t)$, where $m(t)$ denotes the mean function of $f(t)$ and $\tilde{f}(t)$ a centered GP. Then, 
\begin{equation}\label{lefirst}
	\begin{split}
		&\meanof{\left(\sup_{t\in T}|f(t)|\right)^2}  \\
		&=\meanof{\left(\sup_{t\in T}|\tilde{f}(t)+m(t)|\right)^2} \\
		&\leq \meanof{\left(\sup_{t\in T}|\tilde{f}(t)|+\sup_{t\in T}|m(t)|\right)^2} \\
		&\leq \meanof{2\left(\sup_{t\in T}|\tilde{f}(t)|\right)^2+2\left(\sup_{t\in T}|m(t)|\right)^2} \\
		&= \meanof{2\left(\sup_{t\in T}|\tilde{f}(t)|\right)^2}+2\left(\sup_{t\in T}|m(t)|\right)^2 \\
		&= 2\meanof{\left(\sup_{t\in T}|\tilde{f}(t)|\right)^2} + 2M^2
	\end{split}
\end{equation}

	To bound $\meanof{\left(\sup_{t\in T}|\tilde{f}(t)|\right)^2}$, we introduce $\tilde{g}(t)=\tilde{f}(t)-\tilde{f}(t_0)$ and choose $t_0=\argmin_{t\in T}\Var(f(t_0))$. Notice that $\tilde{g}(t)$ is centered, too, and $\tilde{g}(t_0)=0$. We have

\begin{equation}\label{lesecond}
	\begin{split}
		&\meanof{\left(\sup_{t\in T}|\tilde{f}(t)|\right)^2} \\
		&=\meanof{\left(\sup_{t\in T}|\tilde{g}(t)+\tilde{f}(t_0)|\right)^2} \\
		&\leq\meanof{\left(\sup_{t\in T}|\tilde{g}(t)|+|\tilde{f}(t_0)|\right)^2} \\
		&\leq\meanof{2\left(\sup_{t\in T}|\tilde{g}(t)|\right)^2+2\tilde{f}(t_0)^2} \\
		&=2\meanof{\left(\sup_{t\in T}|\tilde{g}(t)|\right)^2}+2\upsilon,
	\end{split}
\end{equation}
where the rhs. of the last equality comes from $\meanof{\tilde{f}(t_0)^2}=\Var(\tilde{f}(t_0))=\Var(f(t_0))=\min_{t\in T}\Var(f(t))=\upsilon$. Thus, we are left with bounding $\meanof{\left(\sup_{t\in T}|\tilde{g}(t)|\right)^2}$. 

We introduce $\tilde{g}^+(t)=max(0,\tilde{g}(t))$ and $\tilde{g}^-(t)=max(0,-\tilde{g}(t))$. Notice that, by symmetry of $\tilde{g}(t)$ around 0, both $\tilde{g}^+(t)$ and $\tilde{g}^-(t)$ must follow the same probability law. We now have
	\begin{equation}\label{lethird}
	\begin{split}
		&\meanof{\left(\sup_{t\in T}|\tilde{g}(t)|\right)^2} \\
		&=\meanof{\left(\sup_{t\in T}\tilde{g}^+(t)+\tilde{g}^-(t)\right)^2} \\
		&\leq\meanof{\left(\sup_{t\in T}\tilde{g}^+(t)+\sup_{t\in T}\tilde{g}^-(t)\right)^2} \\
		&\leq\meanof{2\left(\sup_{t\in T}\tilde{g}^+(t)\right)^2+2\left(\sup_{t\in T}\tilde{g}^-(t)\right)^2} \\
		&=4\meanof{\left(\sup_{t\in T}\tilde{g}^+(t)\right)^2} \\
		&=4\meanof{\left(\sup_{t\in T}\tilde{g}(t)\right)^2} \\
		&=4\meanof{\left(\sup_{t\in T}\tilde{g}(t)\right)}^2+4\Var\left(\sup_{t\in T}\tilde{g}(t)\right).
	\end{split}
\end{equation}
	The second last equality follows from the fact $\tilde{g}^+(t_0)=\tilde{g}(t_0)$, so $\sup_{t\in T}\tilde{g}^+(t)=\sup_{t\in T}\tilde{g}(t)$. Since $\tilde{g}(\cdot)$ is centered, we can apply the Borell-TIS Theorem as in \cite{adler2010geometry}. The boundedness condition for $\tilde{g}(t)$ is fullfilled by the required finite countability of $T$. An immediate consequence is then $\meanof{\left(\sup_{t\in T}\tilde{g}(t)\right)}=K<\infty$. Thus, we are left with finding a bound for $\Var\left(\sup_{t\in T}\tilde{g}(t)\right)$.

Define
\begin{equation}
	\sigma_T^2=\sup_{t\in T}\meanof{\tilde{g}^2(t)}.
\end{equation}
By the Borel-TIS Theorem, it follows that
\begin{equation}
	P\left(\sup_{t\in T}\tilde{g}(t)-\meanof{\sup_{t\in T}\tilde{g}(t)}>u\right)\leq\expof{\frac{-u^2}{2\sigma_T^2}}.
\end{equation}
Due to symmetry around 0 for each $\tilde{g}(t)$, we also have 
\begin{equation}
	\begin{split}
		&P\left(\sup_{t\in T}\tilde{g}(t)-\meanof{\sup_{t\in T}\tilde{g}(t)}<-u\right) \\
		&\leq\expof{\frac{-u^2}{2\sigma_T^2}}
	\end{split}
\end{equation}
and therefore
\begin{equation}
	\begin{split}
		&P\left(\big|\sup_{t\in T}\tilde{g}(t)-\meanof{\sup_{t\in T}\tilde{g}(t)}\big|>u\right) \\
		&P\left(\sup_{t\in T}\tilde{g}(t)-\meanof{\sup_{t\in T}\tilde{g}(t)}>u\right)+P\left(\sup_{t\in T}\tilde{g}(t)-\meanof{\sup_{t\in T}\tilde{g}(t)}<-u\right)\\
		&\leq2\expof{\frac{-u^2}{2\sigma_T^2}}.
	\end{split}
\end{equation}
Finally, 
\begin{equation}\label{lefourth}
	\begin{split}
		&\Var\left(\sup_{t\in T}\tilde{g}(t)\right) \\
		&=\meanof{\bigg|\sup_{t\in T}\tilde{g}(t)-\meanof{\sup_{t\in T}\tilde{g}(t)}\bigg|^2} \\
		&=\int_0^\infty 2uP\left(\bigg|\sup_{t\in T}\tilde{g}(t)-\meanof{\sup_{t\in T}\tilde{g}(t)}\bigg|>u\right)du \\
		&\leq \int_0^\infty 2u 2\expof{\frac{-u^2}{2\sigma_T^2}}du \\
		&= 4\sigma^2_T
	\end{split}
\end{equation}
Putting \eqref{lefirst}, \eqref{lesecond}, \eqref{lethird}, \eqref{lefourth} together, we get
\begin{equation}
		\meanof{\left(\sup_{t\in T}|f(t)|\right)^2}\leq 64\sigma^2_T+16K^2+4v+2M^2 < \infty
\end{equation}
\end{proof}

\begin{lemma}
	Let $H:\mathbb{R}^W \mapsto  \mathbb{R}^C$ denote an almost-everywhere continuous function such that, for $\faty\in\mathbb{R}^C$, either
	\begin{itemize}
		\item $\sup_{j\in 1,...,C}\left(|H(\faty)|\right)_{(j)}\leq D\left(1+\left(\sum_{j=1}^W|\faty_{(j)}|\right)^r\right),\quad D>0,r\geq 0$ or
		\item $\sup_{j\in 1,...,C}\left(|H(\faty)|\right)_{(j)}\leq \kappa\expof{\sum_{(j)=1}^W\gamma_j \fatx_{(j)}},\quad \kappa>0, \gamma_j\geq0$
	\end{itemize}
	Let $F:\mathbb{R}^M\mapsto\mathbb{R}^W, \tilde{F}:\mathbb{R}^M\mapsto\mathbb{R}^C$ denote two vector-valued GPs with arbitrary inter-dependency structure. We then have
	\begin{equation}
		\int_0^1\meanof{\absof{H(F(\fatx_{\alpha}))^T \cdot \tilde{F}(\fatx_\alpha)}}d\alpha<\infty
	\end{equation}
\end{lemma}

\begin{proof}
\begin{equation}
	\begin{split}
		&\int_0^1\meanof{\absof{H(F(\fatx_{\alpha}))^T \cdot \tilde{F}(\fatx_\alpha)}}d\alpha \\
		&=\int_0^1\meanof{\absof{\sum_{j=1}^C H(F(\fatx_{\alpha}))_{(j)} \cdot \tilde{F}(\fatx_\alpha)_{(j)}}}d\alpha \\
		&\leq\int_0^1\meanof{\sum_{j=1}^C\absof{ H(F(\fatx_{\alpha}))_{(j)} \cdot \tilde{F}(\fatx_\alpha)_{(j)}}}d\alpha \\
		&\leq\int_0^1\meanof{C\sup_{j\in 1,...,C}\absof{ H(F(\fatx_{\alpha}))_{(j)} \cdot \tilde{F}(\fatx_\alpha)_{(j)}}}d\alpha \\
		&\leq C\int_0^1\meanof{\sup_{j\in 1,...,C}\absof{ H(F(\fatx_{\alpha}))_{(j)}} \cdot \sup_{j\in 1,...,C}\absof{\tilde{F}(\fatx_\alpha)_{(j)}}}d\alpha \\
		&\leq C\int_0^1\sqrt{\meanof{\squareof{\sup_{j\in 1,...,C}\absof{ H(F(\fatx_{\alpha}))_{(j)}}}}} \cdot \sqrt{\meanof{\squareof{\sup_{j\in 1,...,C}\absof{\tilde{F}(\fatx_\alpha)_{(j)}}}}}d\alpha \\
	\end{split}
\end{equation}
	Denote now $W_1\hat{=}\meanof{\squareof{\sup_{j\in 1,...,C}\absof{ H(F(\fatx_{\alpha}))_{(j)}}}}$. Then $W_1<\infty$ either by Lemma 4 or by the finite expectation of the $\exp$ of a Gaussian random variable. By Lemma 5, we also have $\meanof{\squareof{\sup_{j\in 1,...,C}\absof{\tilde{F}(\fatx_\alpha)_{(j)}}}}\hat{=}W_2<\infty$ and therefore:
\begin{equation}
	\begin{split}
		&\int_0^1\meanof{\absof{H(F(\fatx_{\alpha}))^T \cdot \tilde{F}(\fatx_\alpha)}}d\alpha \\
		&\leq C\int_0^1\sqrt{W_1}\sqrt{W_2}d\alpha \\
		&=C\sqrt{W_1}\sqrt{W_2} <\infty,
	\end{split}
\end{equation}
which concludes the proof.
\end{proof}
To match notation in Lemma 6, we replace $H$ by either $\nabla G_c$ or $g'$ and $\tilde{F}$ by either $F^{\Jk}$ or $f^{\partial_k}$. $F$ is either a vector-valued GP and notation stays the same or single-valued, which we denote again as $f$. Notice that the bounds we placed on $H$ apply the gradient and derivatives of all proposed inverse link-functions.
\end{proof}


\section{E. Experiment Details}
\subsection{E.1. Details on the synthetic data experiment}
All synthetic datasets are sampled from the following distributions:
\begin{equation}
p(\faty|\fatx;\fatW)=p(\faty|H(\fatW\cdot\sin(2\fatx))),\quad p(\fatx)=\mathcal{U}(\fatx|-2,2),\quad p(\veco\fatW)=\mathcal{N}(\veco\fatW|\fatzero,\fatI/5)
\end{equation}
Where $\fatW\in\mathbb{R}^{\text{out}\times\text{5}}$, $H$ denotes either a univariate or multivariate inverse link-function and $\veco$ the vectorization operator. The exact target distributions for each model are:
\begin{itemize}
	\item $p(y|\fatx;\fatW)=\mathcal{N}\left(0,\left(\fatW\sin(2\fatx)\right)^2\right); W\in\mathbb{R}^{1\times5}$ 
	\item $p(y|\fatx;\fatW)=\mathcal{P}oi\left(\expof{\fatW\sin(2\fatx)}\right); W\in\mathbb{R}^{1\times5}$ 
	\item $p(y|\fatx;\fatW)=\mathcal{B}er\left(\Phi\left(\fatW\sin(2\fatx)\right)\right); W\in\mathbb{R}^{1\times5}$ 
	\item $p(y|\fatx;\fatW)=\mathcal{B}er\left(s\left(\fatW\sin(2\fatx)\right)\right); W\in\mathbb{R}^{1\times5}$ 
	\item $p(y|\fatx;\fatW)=\mathcal{C}at\left(S\left(\fatW\sin(2\fatx)\right)\right); W\in\mathbb{R}^{5\times5}$ 
\end{itemize}
We then fit an SVGP model with corresponding target distribution and 50 inducing points (initialized to be the first 50 training points) on a sample dataset consisting of 500 datapoints from the respective distribution.

Next, the Integrated Gradients are calculated on 50 random points from the dataset and their sum is compared against the model output difference for the target point and an all-zero baseline. The average of these absolute difference for each model and number of Riemann interpolation points are then stated in Figure \ref{riemannsumtable}.

\subsection{E.2. MNIST/FashionMNIST Ablation experiment - motivation and rationale}
The rationale is as follows - all images are presumed to be color-encoded, where 2D grayscale or 3D RGB images are flattened into vector-representation: Consider a target image $\fatx$, a baseline image $\fatx^B=\fatzero$ and a perturbation of the target image, denoted as $\tilde{fatx}$. Here, the perturbations are generated by replacing the $k$-th pixel from the target image by the respective value from the baseline image:
\begin{equation}
	\tilde{\fatx}_{(k)}=\fatx^B_{(k)}=0
\end{equation}
If the proposed method produces coherent attributions, replacing a pixel or feature with positive attribution by its baseline value should lead to a decrease in class probability.
\begin{equation}
	\begin{split}
		&\meanof{S_c(F(\fatx))}-\meanof{S_c(F(\fatx^B))}>\meanof{S_c(F(\tilde{\fatx}))}-\meanof{S_c(F(\fatx^B))} \\
		&\Leftrightarrow \meanof{S_c(F(\fatx))}-\meanof{S_c(F(\tilde{\fatx}))}>0
	\end{split}
\end{equation}
The opposite effect is expected when dropping a feature with negative attribution. As we are removing more and more features with positive attribution from the target image, the output difference between original target and perturbated target should keep increasing; the reverse should again hold for removing features with negative attribution. 

Regarding the size of the attributions
\begin{equation}
	\begin{split}
		&\hat{IG}(\fatx,\fxb)_a^{S_c\circ F}>\hat{IG}(\fatx,\fxb)_b^{S_c\circ F}>0 \\
		&\Rightarrow \meanof{S_c(F(\fatx))}-\meanof{S_c(F(\fatx^B))}>\meanof{S_c(F(\tilde{\fatx}^b))}-\meanof{S_c(F(\fatx^B))}>\meanof{S_c(F(\tilde{\fatx}^a))}-\meanof{S_c(F(\fatx^B))},
	\end{split}
\end{equation}
with, again, opposite effect for negative attribution. 
As a consequence, replacing pixels from the target image by their baseline, in order of the corresponding feature attribution from largest to smallest, and keeping pixels with zero attribution should result in the following observation:

As we are drop pixels with large positive attribution, $\meanof{S_c(F(\fatx))}-\meanof{S_c(F(\tilde{\fatx}))}$ should increase steeply. Once features with attribution only slightly \emph{above} zero are replaced, the increase should become flatter. When features with attribution slightly \emph{below} zero are removed, we should see a slight decrease at first until pixels with large negative attribution are remove. At this point, we can expect to see a sharper drop until 
\begin{equation}
\meanof{S_c(F(\fatx))}-\meanof{S_c(F(\tilde{\fatx}))} \equiv \meanof{S_c(F(\fatx))}-\meanof{S_c(F(\fatx^B))},
\end{equation}
at which point the difference should be close to the sum of Integrated Gradients for the original target (as predicted in Theorem 2).

\subsection{E.3. MNIST/FashionMNIST Ablation experiment - further examples}
\begin{figure}[htbp!]
  \centering
  \begin{subfigure}[t]{.45\linewidth}
    \centering
    \begin{minipage}{.5\linewidth}
		\raisebox{0.45cm}{
		  \includegraphics[width=\linewidth]{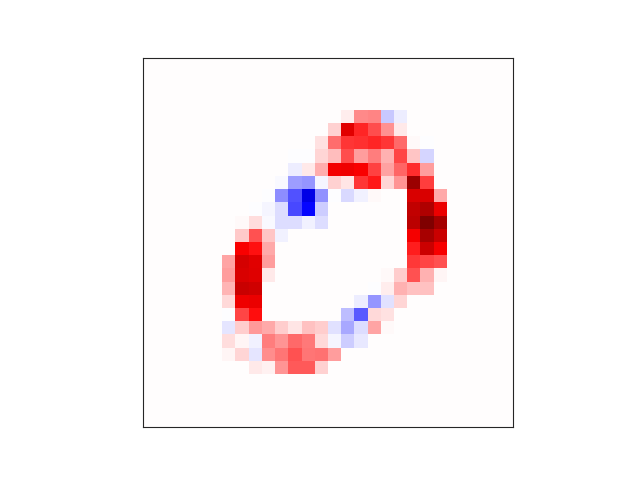}
		}
    \end{minipage}%
    \begin{minipage}{.5\linewidth}
      \includegraphics[width=\linewidth]{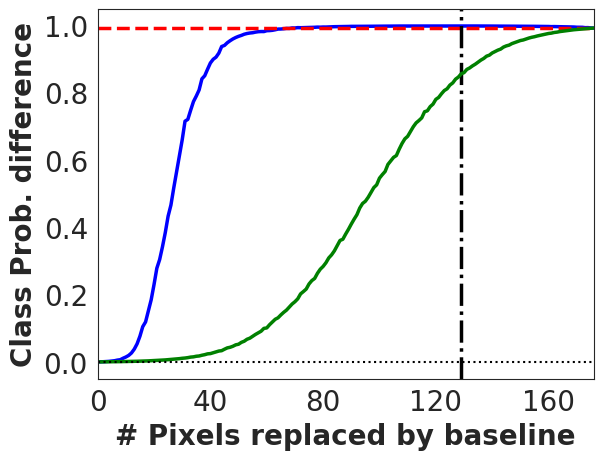}
    \end{minipage}
  \end{subfigure}
  \begin{subfigure}[t]{.45\linewidth}
    \centering
    \begin{minipage}{.5\linewidth}
		\raisebox{0.45cm}{
		  \includegraphics[width=\linewidth]{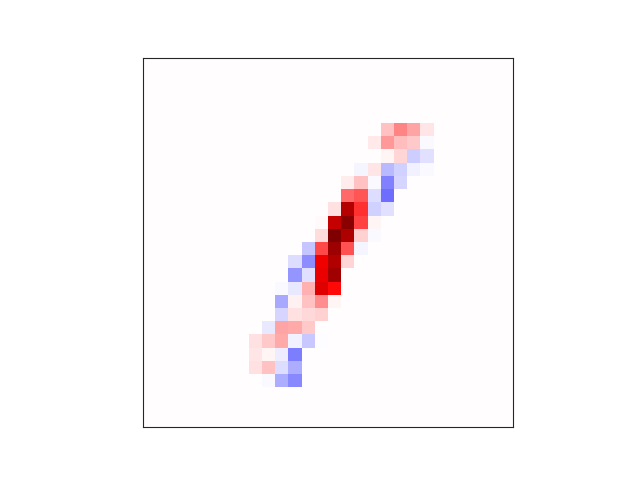}
		}
    \end{minipage}%
    \begin{minipage}{.5\linewidth}
      \includegraphics[width=\linewidth]{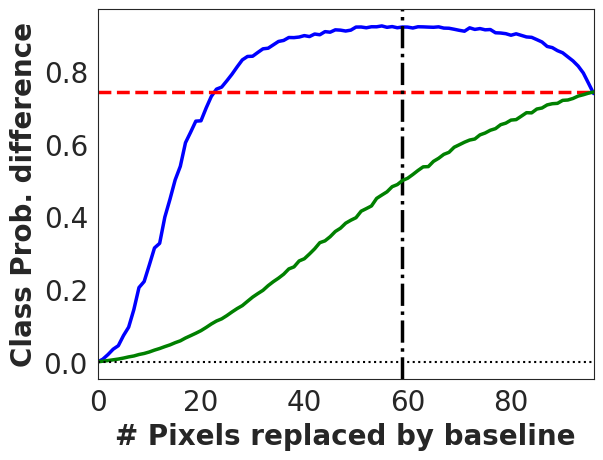}
    \end{minipage}
  \end{subfigure}
  \medskip
  \begin{subfigure}[t]{.45\linewidth}
    \centering
    \begin{minipage}{.5\linewidth}
		\raisebox{0.45cm}{
		  \includegraphics[width=\linewidth]{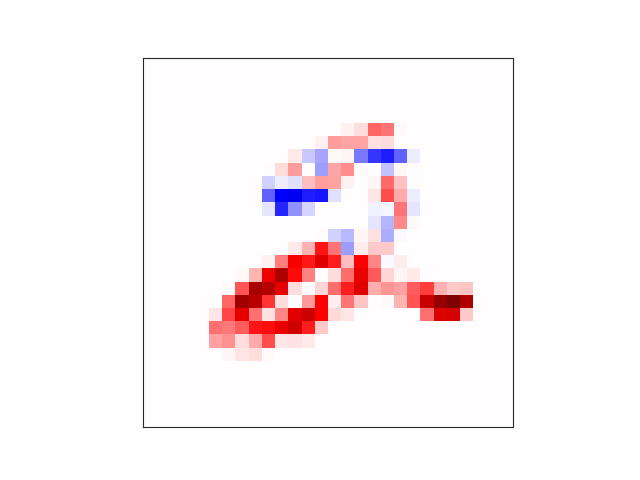}
		}
    \end{minipage}%
    \begin{minipage}{.5\linewidth}
      \includegraphics[width=\linewidth]{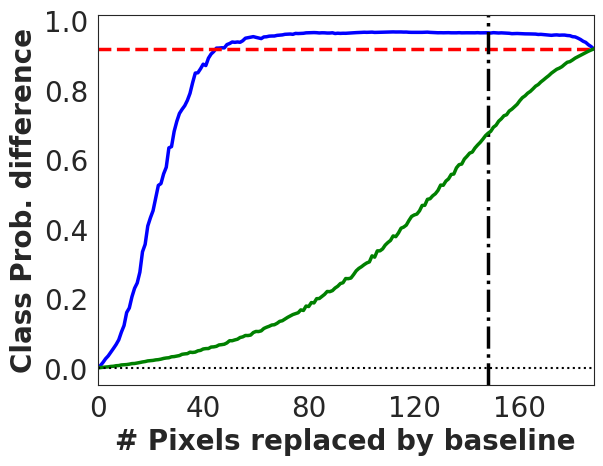}
    \end{minipage}
  \end{subfigure}
  \begin{subfigure}[t]{.45\linewidth}
    \centering
    \begin{minipage}{.5\linewidth}
		\raisebox{0.45cm}{
		  \includegraphics[width=\linewidth]{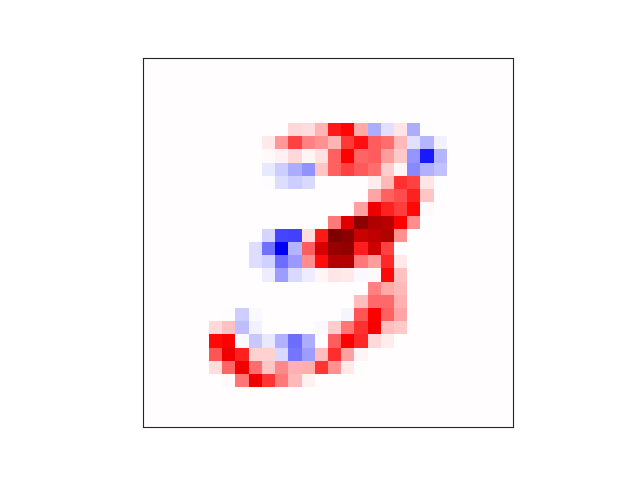}
		}
    \end{minipage}%
    \begin{minipage}{.5\linewidth}
      \includegraphics[width=\linewidth]{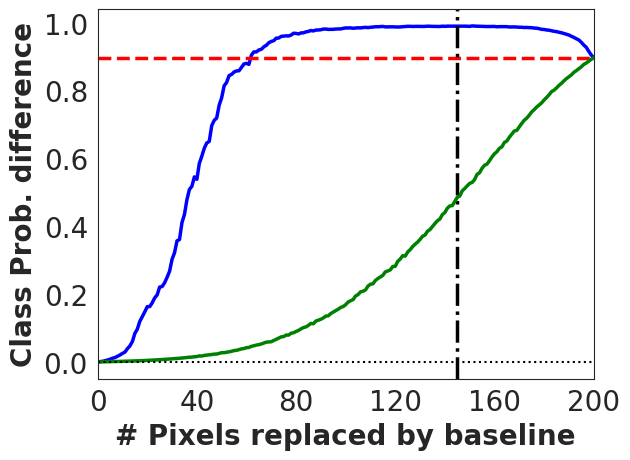}
    \end{minipage}
  \end{subfigure}
  \medskip
  \begin{subfigure}[t]{.45\linewidth}
    \centering
    \begin{minipage}{.5\linewidth}
		\raisebox{0.45cm}{
		  \includegraphics[width=\linewidth]{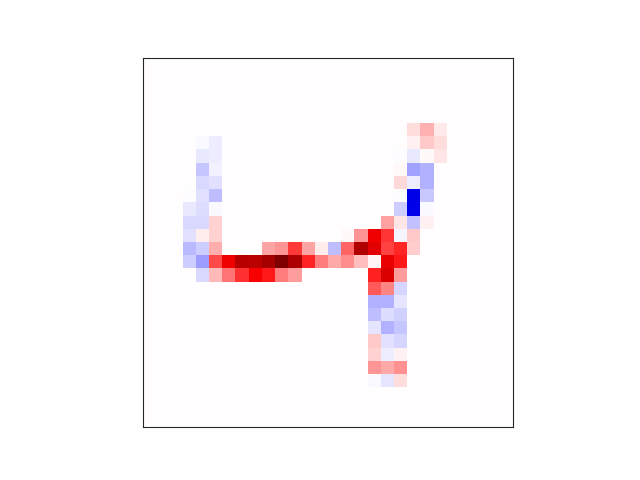}
		}
    \end{minipage}%
    \begin{minipage}{.5\linewidth}
      \includegraphics[width=\linewidth]{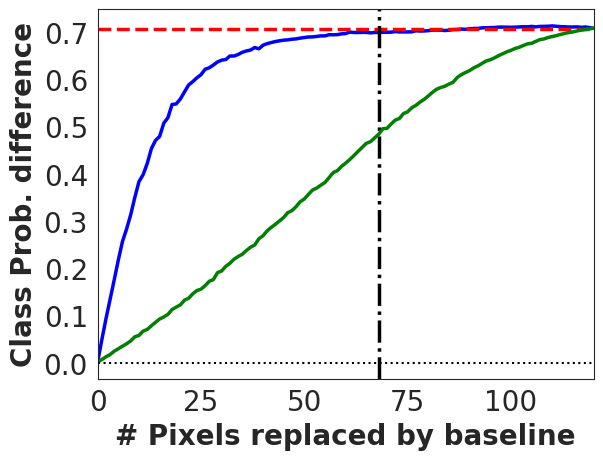}
    \end{minipage}
  \end{subfigure}
  \begin{subfigure}[t]{.45\linewidth}
    \centering
    \begin{minipage}{.5\linewidth}
		\raisebox{0.45cm}{
		  \includegraphics[width=\linewidth]{img/experiments/mnist_5_image.png}
		}
    \end{minipage}%
    \begin{minipage}{.5\linewidth}
      \includegraphics[width=\linewidth]{img/experiments/mnist_5_chart.png}
    \end{minipage}
  \end{subfigure}
  \medskip
  \begin{subfigure}[t]{.45\linewidth}
    \centering
    \begin{minipage}{.5\linewidth}
		\raisebox{0.45cm}{
		  \includegraphics[width=\linewidth]{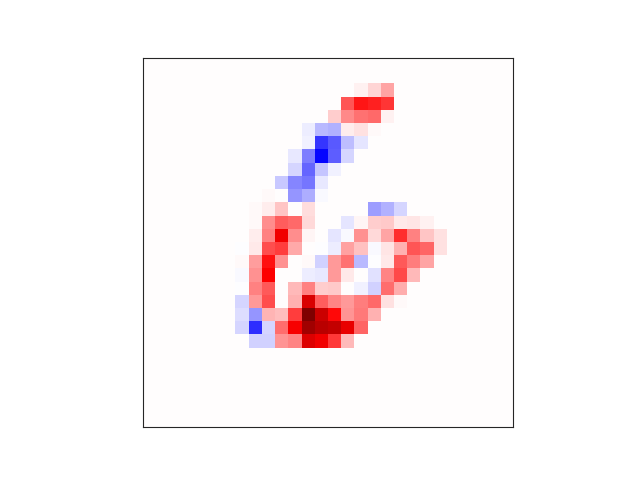}
		}
    \end{minipage}%
    \begin{minipage}{.5\linewidth}
      \includegraphics[width=\linewidth]{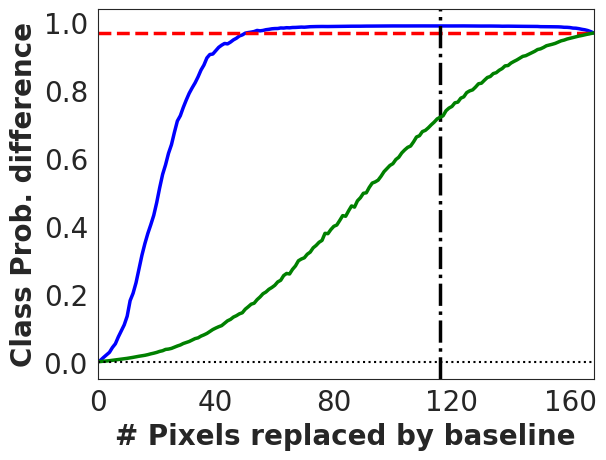}
    \end{minipage}
  \end{subfigure}
  \begin{subfigure}[t]{.45\linewidth}
    \centering
    \begin{minipage}{.5\linewidth}
		\raisebox{0.45cm}{
		  \includegraphics[width=\linewidth]{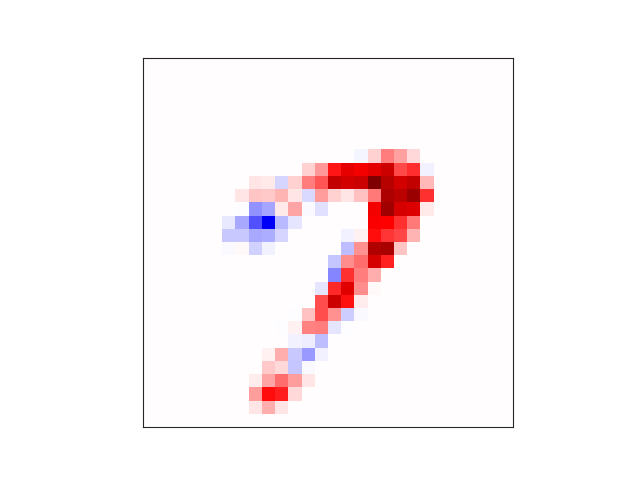}
		}
    \end{minipage}%
    \begin{minipage}{.5\linewidth}
      \includegraphics[width=\linewidth]{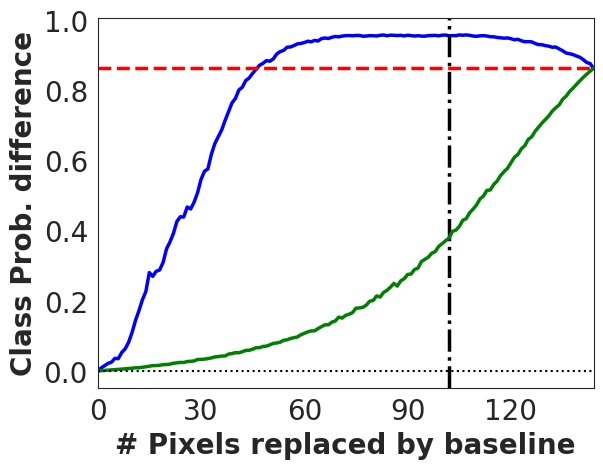}
    \end{minipage}
  \end{subfigure}
  \medskip
  \begin{subfigure}[t]{.45\linewidth}
    \centering
    \begin{minipage}{.5\linewidth}
		\raisebox{0.45cm}{
		  \includegraphics[width=\linewidth]{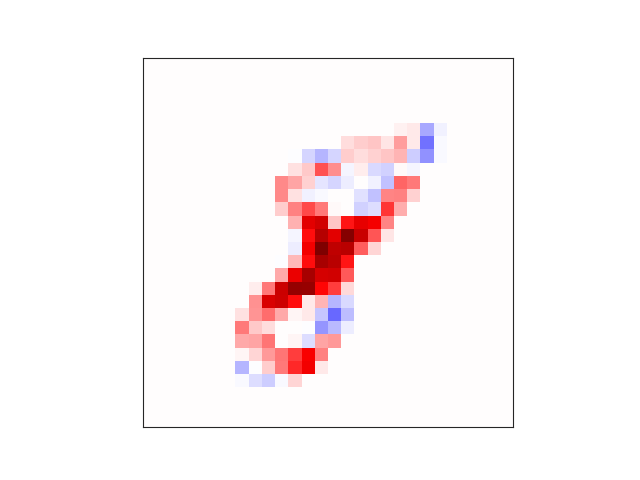}
		}
    \end{minipage}%
    \begin{minipage}{.5\linewidth}
      \includegraphics[width=\linewidth]{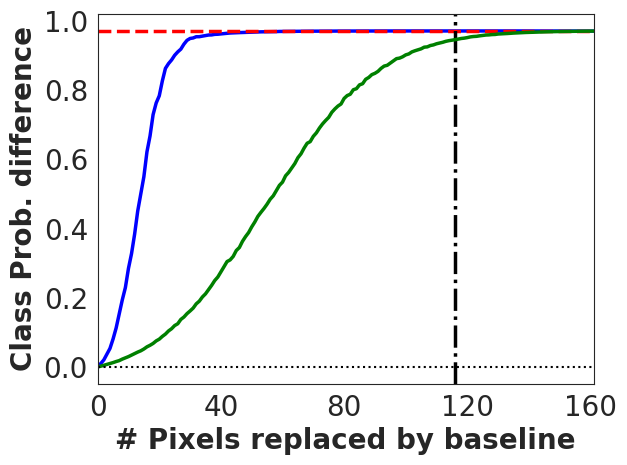}
    \end{minipage}
  \end{subfigure}
  \begin{subfigure}[t]{.45\linewidth}
    \centering
    \begin{minipage}{.5\linewidth}
		\raisebox{0.45cm}{
		  \includegraphics[width=\linewidth]{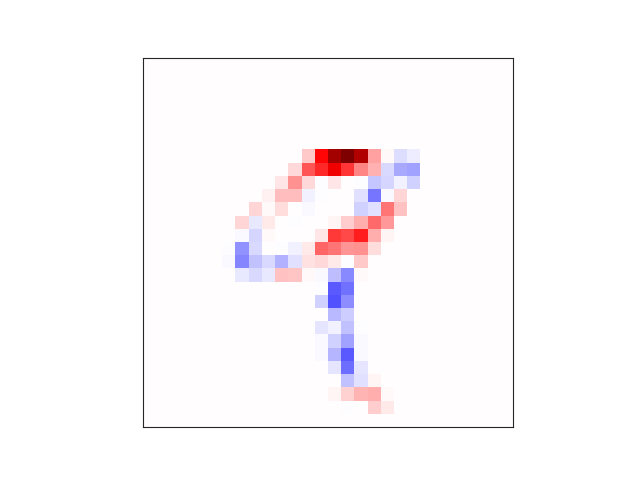}
		}
    \end{minipage}%
    \begin{minipage}{.5\linewidth}
      \includegraphics[width=\linewidth]{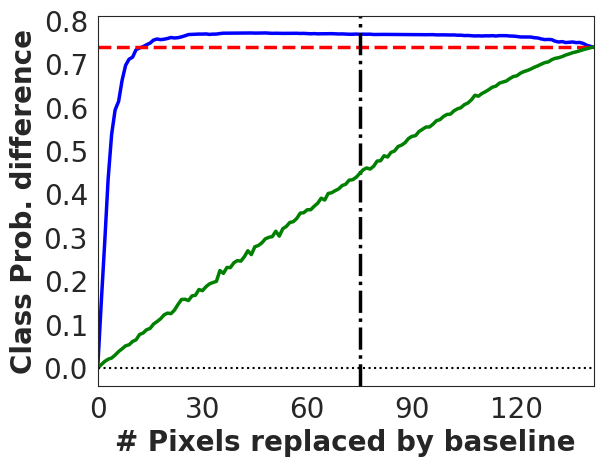}
    \end{minipage}
  \end{subfigure}
  \medskip
  \label{fig:mnist_plot_all}
  \caption{\textit{Further results from the ablation experiments on MNIST}}
\end{figure}

\begin{figure}[htbp!]
  \centering
  \begin{subfigure}[t]{.45\linewidth}
    \centering
    \begin{minipage}{.5\linewidth}
		\raisebox{0.45cm}{
		  \includegraphics[width=\linewidth]{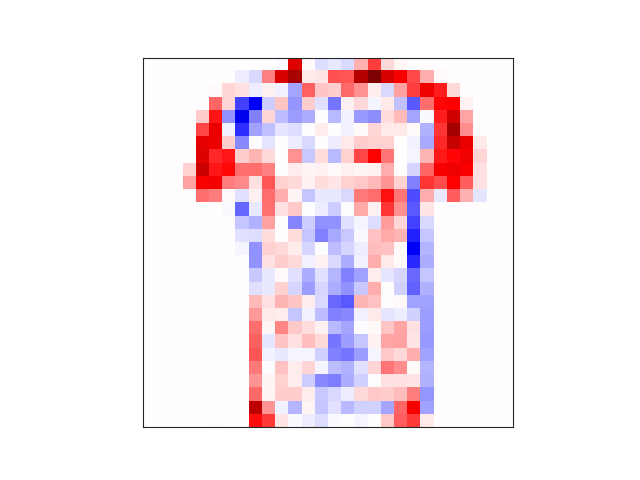}
		}
    \end{minipage}%
    \begin{minipage}{.5\linewidth}
      \includegraphics[width=\linewidth]{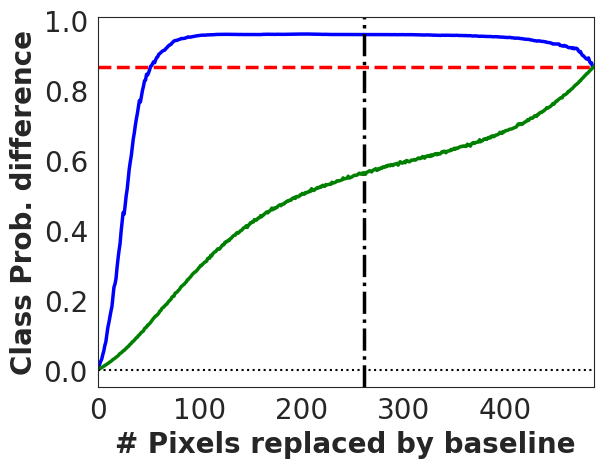}
    \end{minipage}
  \end{subfigure}
  \begin{subfigure}[t]{.45\linewidth}
    \centering
    \begin{minipage}{.5\linewidth}
		\raisebox{0.45cm}{
		  \includegraphics[width=\linewidth]{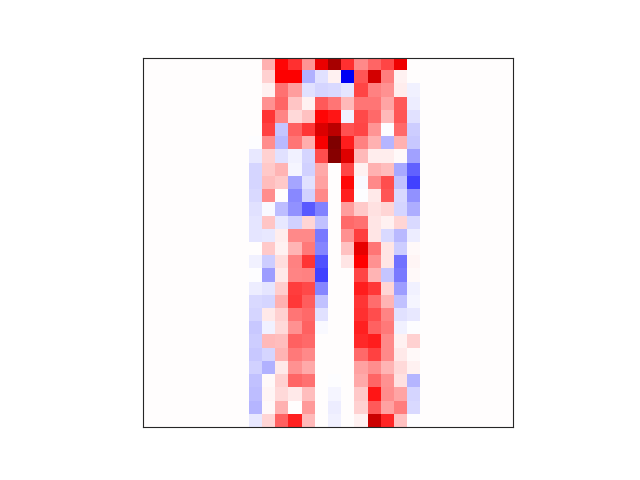}
		}
    \end{minipage}%
    \begin{minipage}{.5\linewidth}
      \includegraphics[width=\linewidth]{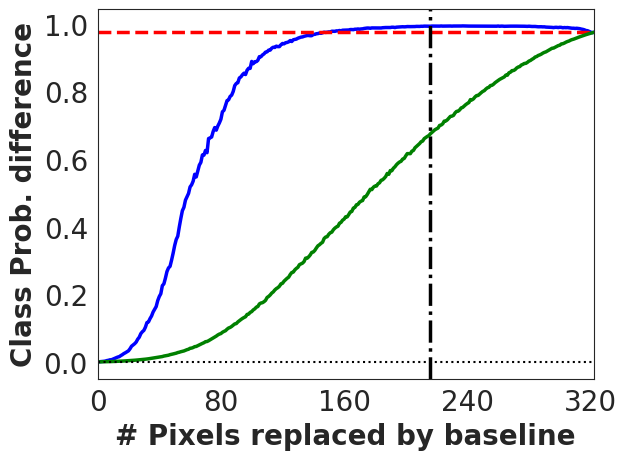}
    \end{minipage}
  \end{subfigure}
  \medskip
  \begin{subfigure}[t]{.45\linewidth}
    \centering
    \begin{minipage}{.5\linewidth}
		\raisebox{0.45cm}{
		  \includegraphics[width=\linewidth]{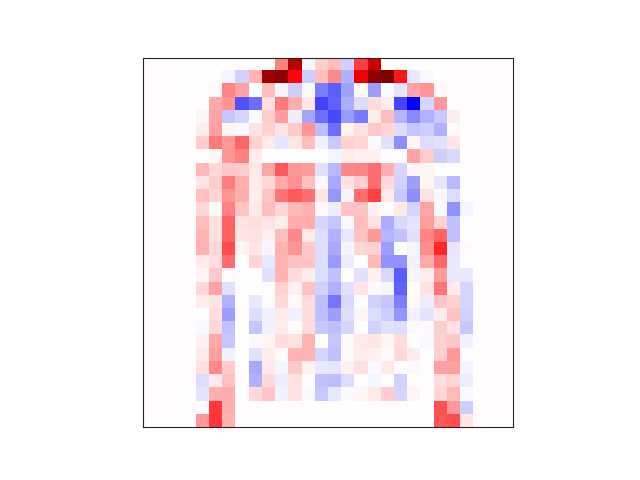}
		}
    \end{minipage}%
    \begin{minipage}{.5\linewidth}
      \includegraphics[width=\linewidth]{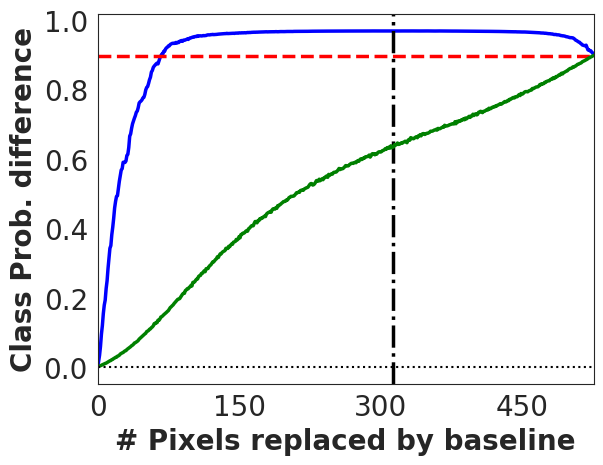}
    \end{minipage}
  \end{subfigure}
  \begin{subfigure}[t]{.45\linewidth}
    \centering
    \begin{minipage}{.5\linewidth}
		\raisebox{0.45cm}{
		  \includegraphics[width=\linewidth]{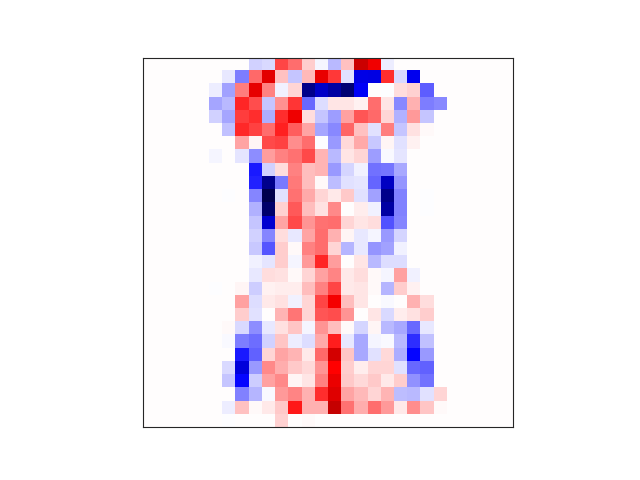}
		}
    \end{minipage}%
    \begin{minipage}{.5\linewidth}
      \includegraphics[width=\linewidth]{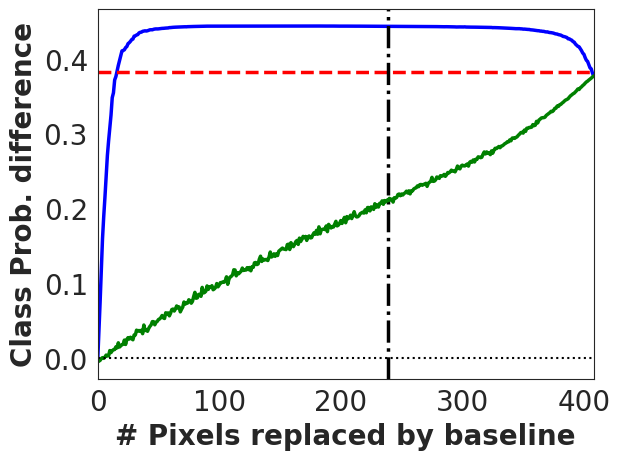}
    \end{minipage}
  \end{subfigure}
  \medskip
  \begin{subfigure}[t]{.45\linewidth}
    \centering
    \begin{minipage}{.5\linewidth}
		\raisebox{0.45cm}{
		  \includegraphics[width=\linewidth]{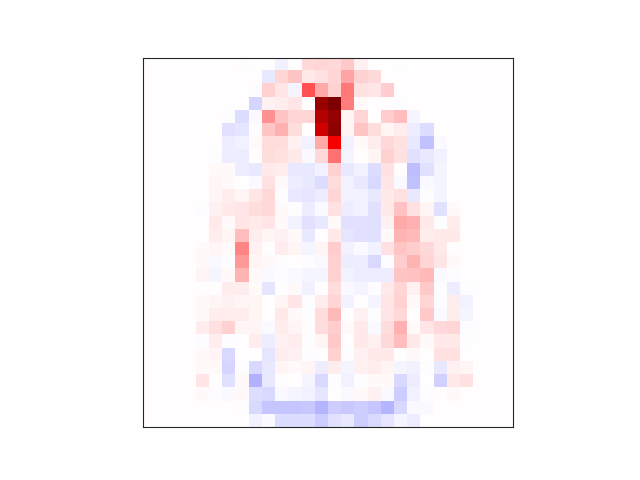}
		}
    \end{minipage}%
    \begin{minipage}{.5\linewidth}
      \includegraphics[width=\linewidth]{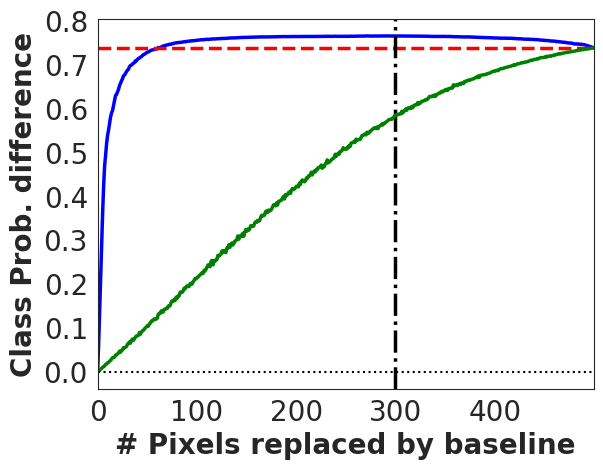}
    \end{minipage}
  \end{subfigure}
  \begin{subfigure}[t]{.45\linewidth}
    \centering
    \begin{minipage}{.5\linewidth}
		\raisebox{0.45cm}{
		  \includegraphics[width=\linewidth]{img/experiments/fashionmnist_5_image.png}
		}
    \end{minipage}%
    \begin{minipage}{.5\linewidth}
      \includegraphics[width=\linewidth]{img/experiments/fashionmnist_5_chart.png}
    \end{minipage}
  \end{subfigure}
  \medskip
  \begin{subfigure}[t]{.45\linewidth}
    \centering
    \begin{minipage}{.5\linewidth}
		\raisebox{0.45cm}{
		  \includegraphics[width=\linewidth]{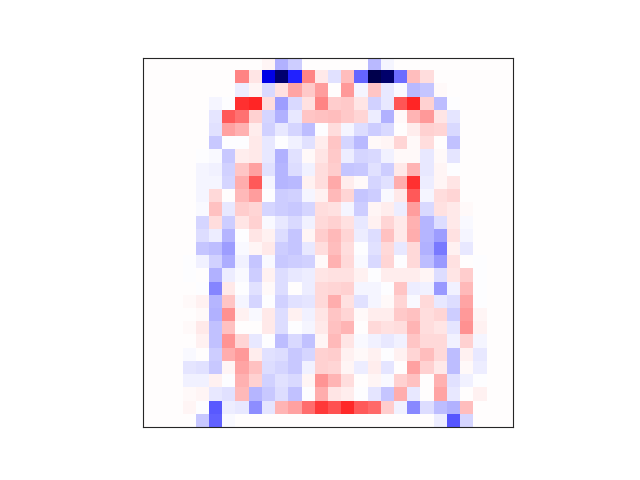}
		}
    \end{minipage}%
    \begin{minipage}{.5\linewidth}
      \includegraphics[width=\linewidth]{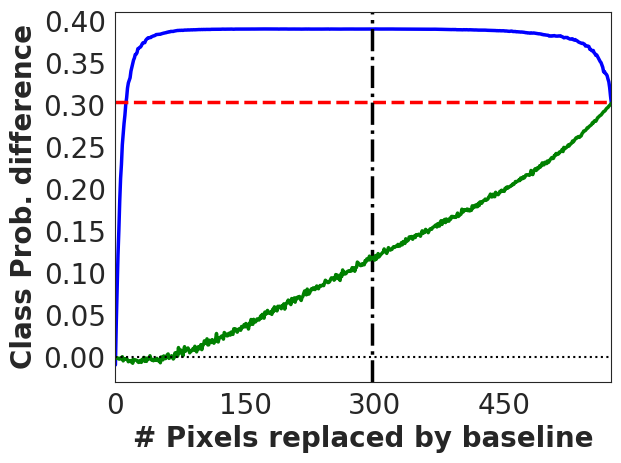}
    \end{minipage}
  \end{subfigure}
  \begin{subfigure}[t]{.45\linewidth}
    \centering
    \begin{minipage}{.5\linewidth}
		\raisebox{0.45cm}{
		  \includegraphics[width=\linewidth]{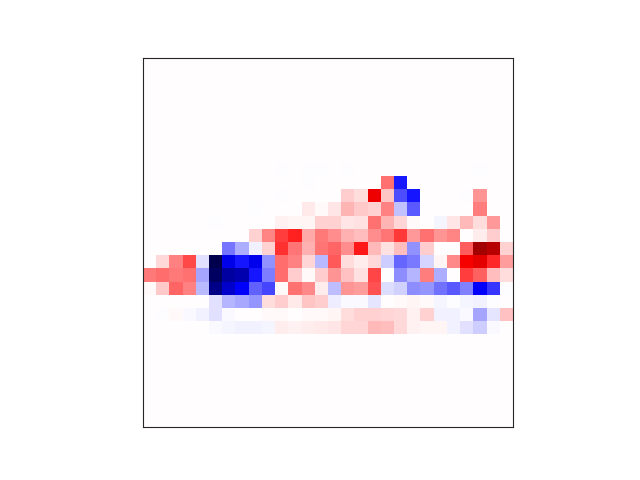}
		}
    \end{minipage}%
    \begin{minipage}{.5\linewidth}
      \includegraphics[width=\linewidth]{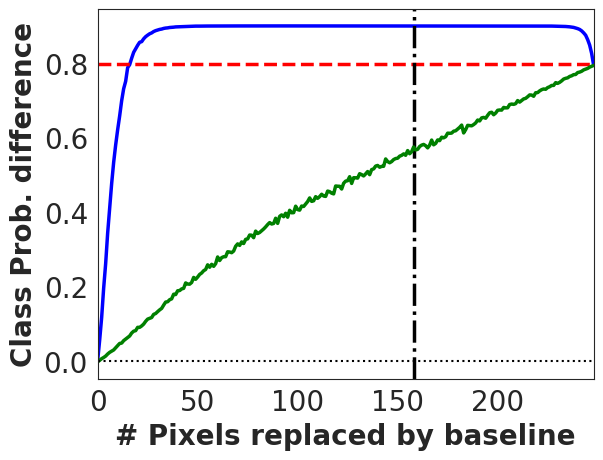}
    \end{minipage}
  \end{subfigure}
  \medskip
  \begin{subfigure}[t]{.45\linewidth}
    \centering
    \begin{minipage}{.5\linewidth}
		\raisebox{0.45cm}{
		  \includegraphics[width=\linewidth]{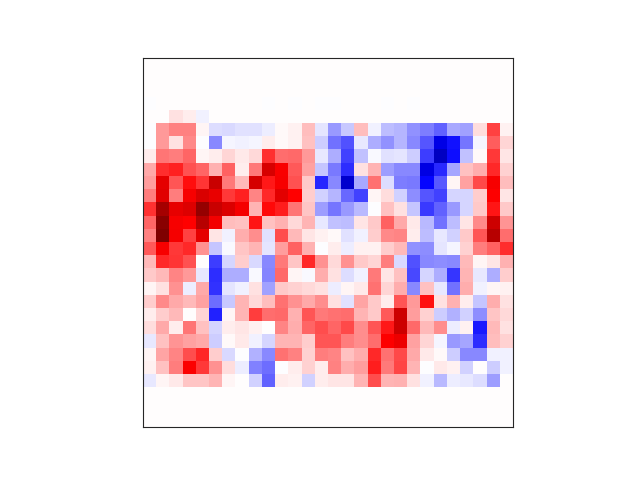}
		}
    \end{minipage}%
    \begin{minipage}{.5\linewidth}
      \includegraphics[width=\linewidth]{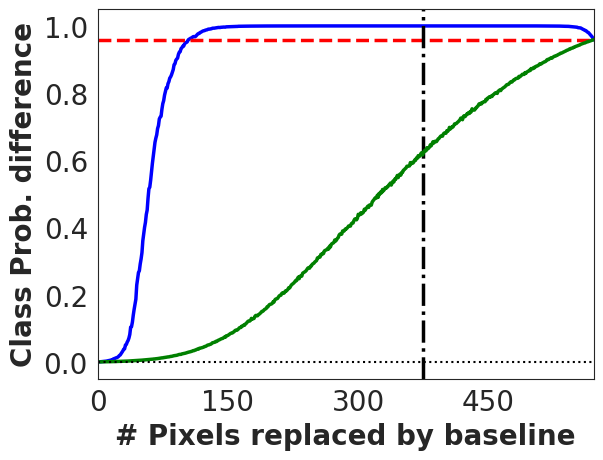}
    \end{minipage}
  \end{subfigure}
  \begin{subfigure}[t]{.45\linewidth}
    \centering
    \begin{minipage}{.5\linewidth}
		\raisebox{0.45cm}{
		  \includegraphics[width=\linewidth]{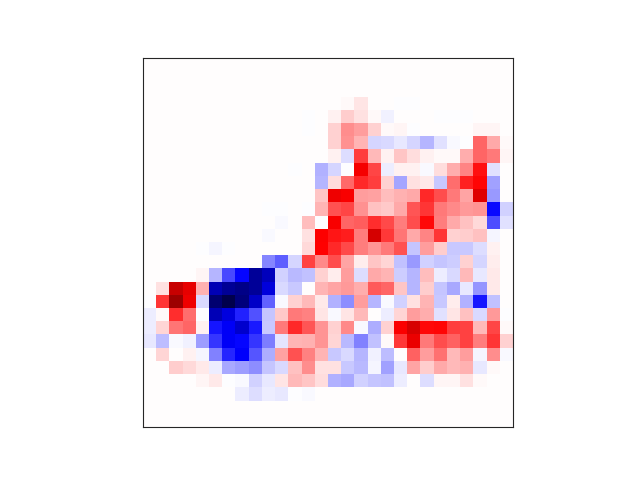}
		}
    \end{minipage}%
    \begin{minipage}{.5\linewidth}
      \includegraphics[width=\linewidth]{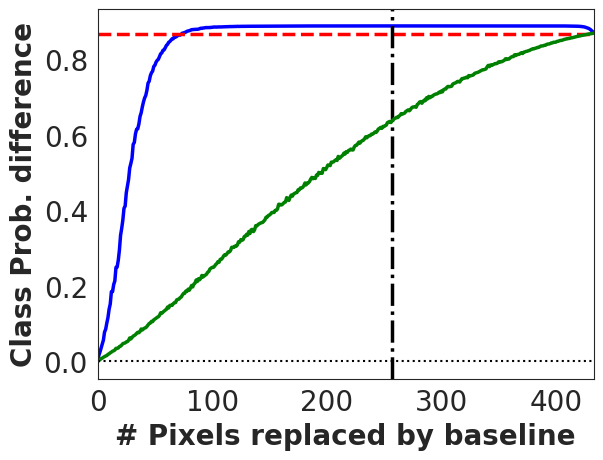}
    \end{minipage}
  \end{subfigure}
  \medskip
  \label{fig:fashionmnist_plot_all}
  \caption{\textit{Further results from the ablation experiments on FashionMNIST}}
\end{figure}

\end{document}